\documentclass[runningheads]{llncs}
\usepackage{graphicx}
\usepackage{comment}
\usepackage{amsmath,amssymb} 
\usepackage{color}

\usepackage[colorlinks = true,
            linkcolor = red,]{hyperref}
\usepackage{caption}
\usepackage{multirow}
\usepackage{xspace}
\usepackage{mathtools}
\usepackage{wasysym}

\usepackage{marvosym}
\usepackage{adjustbox}

\DeclareMathOperator*{\argmin}{arg\,min}

\usepackage{pdfcomment}
\usepackage{lipsum}
\usepackage{algorithm}
\usepackage[noend]{algpseudocode}
\usepackage{wrapfig}

\newcommand{\ourmodel}{\textsc{DefGrid}}
\newcommand{\fixedgrid}{\textsc{FixedGrid}}

\begin{document}
\pagestyle{headings}
\mainmatter
\def\ECCVSubNumber{735}  
\title{Beyond Fixed Grid: Learning Geometric Image Representation with a Deformable Grid}

\titlerunning{Deformable Grid}
%
\author{Jun Gao$^{1,2,3}$ \and
Zian Wang$^{1,2}$ \and
Jinchen Xuan$^{4}$ \and Sanja Fidler$^{1,2,3}$}
\authorrunning{J. Gao et al.}

\institute{$^1$University of Toronto, $^2$Vector Institute,  $^3$NVIDIA, $^4$Peking University  \\
\email{\{jungao, zianwang, fidler\}@cs.toronto.edu},
\email{1600012865@pku.edu.cn}
}

\maketitle

\begin{abstract}
In modern computer vision, images are typically represented as a fixed uniform grid with some stride and processed via a deep convolutional neural network. We argue that deforming the grid to better align with the high-frequency image content is a more effective strategy. We introduce \emph{Deformable Grid} (\ourmodel), a learnable neural network module that predicts location offsets of vertices of a 2-dimensional triangular grid, such that the edges of the deformed grid align with image boundaries. 
We showcase our {\ourmodel} in a variety of use cases, i.e., by inserting it as a module at various levels of processing. 
We utilize {\ourmodel} as an end-to-end \emph{learnable geometric downsampling} layer that replaces standard pooling methods for reducing feature resolution when feeding images into a deep CNN. We show significantly improved results at the same grid resolution compared to using CNNs on uniform grids for the task of semantic segmentation.
We also utilize  {\ourmodel} at the output layers for the task of object mask annotation, and show that reasoning about object boundaries on our predicted polygonal grid leads to more accurate results over existing pixel-wise and curve-based approaches. We finally showcase {\ourmodel} as a standalone module for unsupervised image partitioning, showing superior performance over existing approaches.
Project website: \href{http://www.cs.toronto.edu/~jungao/def-grid}{http://www.cs.toronto.edu/$\sim$jungao/def-grid}.
\end{abstract}

\section{Introduction}

In modern computer vision approaches, an image is treated as a fixed uniform grid with a stride and processed through a deep convolutional neural network. Very high resolution images are typically processed at a lower resolution for increased efficiency, whereby the image is essentially blurred and subsampled. When fed to a neural network, each pixel thus contains a blurry version of the original signal mixing information from both the foreground and background, possibly causing higher sensitivity and reliance  of the network to objects and their context. In contrast, in many of the traditional computer vision pipelines the high resolution image was instead partitioned into a smaller set of superpixels that conform to image boundaries, leading to more effective reasoning in downstream tasks. We follow this line of thought and argue that deforming the grid to better align with the high-frequency information content in the input is a more effective representation strategy. This is conceptually akin to superpixels but conforming to a regular topology with geometric constraints thus still easily amenable for use with deep convolutional networks for downstream tasks. 


\begin{figure}[t!]
\centering
\includegraphics[width=0.9\linewidth, trim=0 12 0 10,clip]{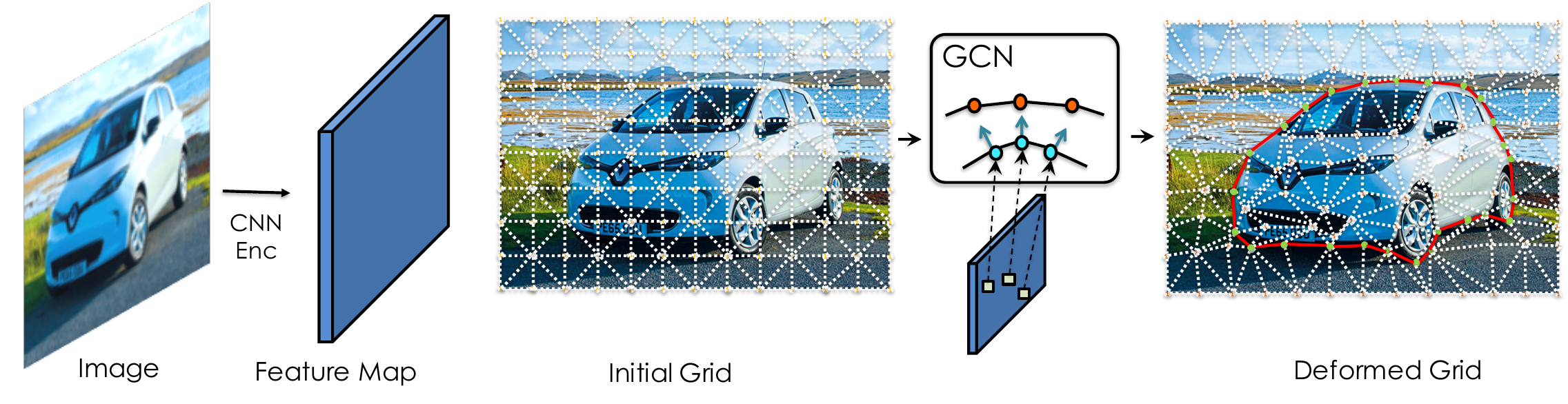}
\caption{{\bf {\ourmodel}} is a neural module that represents an image with a triangular grid. Initialized with an uniform grid, {\ourmodel} deforms grid's vertices, such that grid's edges align with  image boundaries, while keeping topology fixed.}
\label{fig:teaser}
\label{topo}
\end{figure}

Furthermore, tasks such as object mask annotation naturally require the output to be in the form of polygons with a manageable number of control points that a human annotator can edit. Previous work either parametrized the output as a closed curve with a fixed number of control points~\cite{curvegcn} or performed pixel-wise labeling followed by a (non-differentiable) polygonization step~\cite{polytransform,delse,dextr}. In the former, the predicted curves typically better utilize shape priors leading to ``well behaved" predictions, however, the output is inherently limited in the genus and complexity of the shape it is able to represent. In contrast, pixel-wise methods can represent shapes of arbitrary genus, however, typically large input/output resolutions are required to produce accurate labeling around object boundaries. We argue that reasoning on a low-resolution polygonal grid that well aligns with image boundaries combines the advantages of the two approaches. 

We introduce \emph{Deformable Grid} (\ourmodel), a neural network module that represents an image with a 2-dimensional triangular grid. The basic element of the grid is a triangular cell with vertices that place the triangle in the image plane. {\ourmodel} is initialized with a uniform grid and utilizes a neural network that 
predicts location offsets of the triangle vertices such that the edges and vertices of the deformed grid align with image boundaries (Fig~\ref{fig:teaser}). We propose several carefully designed loss functions that encourage this behaviour. 
Due to the differentiability of the deformation operations, {\ourmodel} can be trained end-to-end with downstream neural networks as a plug-and-play module at various levels of deep processing. We showcase {\ourmodel} in various use cases: as a learnable geometric image downsampling layer that affords high accuracy semantic segmentation at significantly reduced grid resolutions. Furthermore, when used to parametrize the output, we show that it leads to more effective and accurate results for the tasks of interactive object mask annotation. Our {\ourmodel} can also be used a standalone module for unsupervised image partitioning, and we show superior performance over existing superpixel-based approaches.

\section{Related Works}
We focus on the most relevant work in several related categories.  

\textbf{Deformable Structure:} Deformable convolutions~\cite{dai2017deformable} predict position offsets of each cell in the convolutional kernel's grid with the aim to better capture object deformation. This is in contrast to our approach which deforms a triangular grid which is then exploited in downstream processing. Note that our approach does not imply any particular downstream neural architecture and would further benefit by employing deformable convolutions. Related to our work,~\cite{recasens2018learning,jaderberg2015spatial} learn to deform an image such that the corresponding warped image, when fed to a neural network, leads to improved downstream tasks. Our {\ourmodel}, which is trained with both unsupervised and supervised loss functions to explicitly align with image boundaries, allows  downstream tasks such as object segmentation to perform reasoning directly on the low-dimensional grid. 

\textbf{Polygonal Image Partitioning:} Polygonal image partitioning plays an important role in certain applications such as multi-view 3D object reconstruction~\cite{bodis2015superpixel} and graphics-based image manipulation~\cite{BBW:2011}. 
Existing works tried to polygonize an image with triangles~\cite{bodis2015superpixel} or convex polygons~\cite{duan2015image}. 
\cite{bodis2015superpixel} construct a 2D triangle mesh over keypoints, traced intensity edges, or polygonized superpixel boundaries, via Constrained Delaunay Triangulation (CDT). However, edge-adherence of such a triangulation assumes precise and sufficiently dense keypoints, accurate edge detection, or superpixels aligned to intensity edges, respectively.~\cite{duan2015image} also relies on the initial line segment detection. In~\cite{snic}, a non-iterative method was proposed to obtain superpixels, followed by a polygonization method using a contour tracing algorithm. 
Our {\ourmodel} produces a triangular grid that conforms to image boundaries and is end-to-end trainable.

\textbf{Superpixels:} 
Superpixel methods aim to partition the image into regions of homogenous color while regularizing their shape and preserving image boundaries~\cite{Felzenszwalb2004,normalizedCut,ers,Grundmann10,Bergh13,slic,snic,turbopixels}. Many algorithms have been proposed mainly differing in the energy function they optimize and the optimization technique they utilize~\cite{normalizedCut,ers,watershed,Felzenszwalb2004,Grundmann10,slic,snic,Yamaguchi14,YaoCVPR15}.  
Most of these approaches produce superpixels with irregular topology, and the final segmentation map is often  disconnected and needs postprocessing.   Note also that energy is often hand designed, and inference is optimization based. 
Recently, SSN~\cite{ssn} made clustering-based approaches differentiable by softly assigning pixels to superpixels with the exponential function. SEAL~\cite{seal} learns superpixels by exploiting segmentation affinities. Both of these methods  produce superpixels with highly irregular boundaries and region topology, and thus they may not be trivially embedded in existing convolutional neural architectures~\cite{gadde2016superpixel}.
To produce regular grid-like topology, superpixel lattices~\cite{Moore08} partition the image recursively, finding horizontal and vertical paths with minimal boundary cost at each iteration. Unlike their approach, {\ourmodel} utilizes differentiable operations to predict boundary aligned triangular grids and is end-to-end trainable with both unsupervised and supervised loss functions. 
\section{Deformable Grid}

\begin{figure}[t!]
    \centering
    \begin{minipage}{0.69\linewidth}
    \centering
    \includegraphics[width=0.92\linewidth]{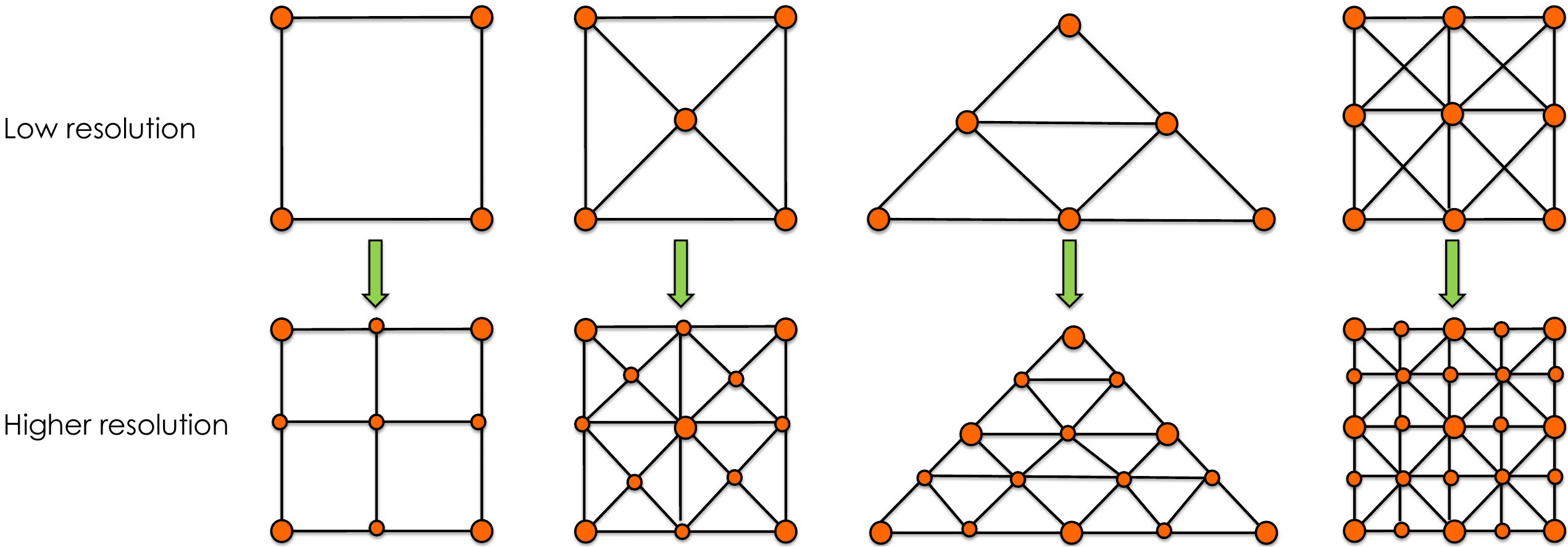}
    \end{minipage}
    \hfill
    \begin{minipage}{0.29\linewidth}
    \caption{{\bf Different grid topologies}. We choose the last column for its flexibility in representing a variety of different edge orientations.}
    \label{fig:grid_topology}
    \end{minipage}
\end{figure}

Our {\ourmodel} is a 2-dimensional triangular grid defined on an image plane. The basic cell in the grid is a triangle with three vertices, each with a location that position the triangle in the image.  
Edges of the triangle thus represent line segments and are expected not to self-intersect across triangles. The topology of the grid is fixed and does not depend on the input image. The geometric grid thus naturally partitions an image into  regular segments, as shown in Fig.~\ref{fig:teaser}.

We formulate our approach as deforming a triangular grid with uniformly initialized vertex positions to better align with image boundaries. The grid is deformed via a neural network that a predicts position offset for each vertex while ensuring the topology does not change (no self-intersections occur). 

Our main intuition is that when the edges of the grid align with image boundaries, the pixels inside each grid cell have minimal variance of RGB values (or one-hot masks when we have supervision),  and vice versa. We  aim to minimize this variance in a  differentiable way with respect to the positions of the vertices, to make it amenable to deep learning. 
We describe our {\ourmodel} formulation along with its training method in detail next. In Section~\ref{sec:applications}, we show applications to different downstream tasks. 


\subsection{Grid Parameterization}

\paragraph{Grid Topology:} Choosing the right topology of the grid is an important aspect of our work. Since objects (and their parts) can appear at different scales in images, we ideally want a topology that can easily be subdivided to accommodate for this diversity. Furthermore, boundaries can be found in any orientation and thus the grid edges should be flexible enough to well align to any real edge. 
We experimentally tried four different topologies which are visualized in Fig.~\ref{fig:grid_topology}. We found the topology in the last column to outperform alternatives for its flexibility in representing different edge orientations. Note that our method is agnostic to the choice of topology and we provide a detailed comparison in the appendix.

\paragraph{Grid Representation:} Let $I$ be an input image. We denote each vertex of the grid in the image plane as $v_i=[x_{i}, y_{i}]^T$, where $i\in\{1,\cdots,n\}$ and $n$ is the total number of vertices in the grid. Since the grid topology is fixed, the grid in the image is entirely specified by the positions of its vertices $\mathbf{v}$. We denote each triangular cell in the grid with its three vertices as $C_k = [v_{a_k}, v_{b_k}, v_{c_k}]$, with $k\in\{1,\cdots,K\}$ indexing the grid cells.  We uniformly initialize the vertices on the 2D image plane, and define {\ourmodel} as a neural network $h$ that  predicts the relative offset for each vertex:
\begin{eqnarray}
\{\Delta x_{i}, \Delta y_{i} \}_{i=1}^{n}= h(\mathbf{v}, I).
\end{eqnarray}
We discuss the choice of $h$ in Section~\ref{sec:applications}. 
The deformed vertices 
are thus:
\begin{eqnarray}
v_{i} &=& [x_{i} + \Delta x_{i}, y_{i} + \Delta y_{i}]^T, \quad\  i=1,\dots,n.
\end{eqnarray}

\subsection{Training of {\ourmodel}} 
We now discuss training of the grid deforming network $h$ using a variety of unsupervised loss functions. We want all our losses to be differentiable with respect to vertex positions to allow for the gradient to be backpropagated analytically.

\paragraph{Differentiable Variance:}
As the grid deforms  (its vertices move), the grid cells will cover different pixel regions in the image. 
Our first loss aims to minimize the variance of pixel features in each grid cell. Each pixel $p_i$ has a feature vector  $\mathbf{f}_i$, which in our case is chosen to contain RGB values. When supervision is available in the form of segmentation masks, we can optionally append a one hot vector representing the class of the mask. Pixel's position in the image is denoted with $p_i=[p_i^x, p_i^y]^T$, $i\in\{1,\cdots,N\}$, with $N$ indicating the total number of pixels in the image. Variance of a cell $C_k$ is defined as:  
\begin{eqnarray}
V_k = \sum\nolimits_{p_i\in S_k} ||\mathbf{f}_i - \overline{\mathbf{f}}_k||^2_2,
\end{eqnarray}
where $S_k$ denotes the set of pixels inside $C_k$, and $\overline{\mathbf{f}}_k$ is the mean feature of $C_k$: $\overline{\mathbf{f}}_k = \frac{\sum_{p_i\in S_k}\mathbf{f}_i}{\sum_{p_i\in S_k} 1}.$
Note that this definition of variance is not naturally differentiable with respect to the vertex positions. We thus reformulate the variance function by softly assigning every pixel $p_i$  to each grid cell $C_k$ with a probability  $P_{i\to k}(\mathbf{v})$:
\begin{eqnarray}
P_{i\to k}(\mathbf{v}) &=& \frac{\exp (\mathrm{SignDis}(p_i, C_k)/\delta)}{\sum_{j=1}^K \exp( \mathrm{SignDis}(p_i, C_j)/\delta)},\\
\mathrm{SignDis}(p_i,C_k) &= &
    \begin{cases}
    -\mathrm{Dis}(p_i,C_k), & \text{if $p_i$ is outside $C_k$,} \\
    \mathrm{Dis}(p_i,C_k), & \text{if $p_i$ is inside $C_k$.}
    \end{cases}\\
\mathrm{Dis}(p_i,C_k) &=& \min(D(p_i,v_{a_k} v_{b_k}),D(p_i,v_{b_k} v_{c_k}), D(p_i,v_{c_k} v_{a_k})), 
\end{eqnarray}
where $D(p_i, v_iv_j)$ is the L1 distance between a pixel and a line segment $v_iv_j$, and $\delta$ is a hyperparameter to control the slackness. We use $P_{i\to k}(\mathbf{v})$ to indicate that the probability of assignment depends on the grid's vertex positions, and is in our case a differentiable function. 
Intuitively, if the pixel is very close or inside a cell, then $P_{i\to k}(\mathbf{v})$ is close to $1$, and close to $0$ otherwise. To check whether the pixel is inside a cell, we calculate the barycentric weight of the pixel with respect to three vertices of the cell. If all the barycentric weights are between 0 and 1, then the pixel is inside, otherwise it falls outside the triangle.

We now re-define the cell's variance as follows:
\begin{eqnarray}
\tilde{V}_k(\mathbf{v}) = \sum\nolimits_{i=1}^N P_{i\to k}(\mathbf{v}) \cdot ||\mathbf{f}_i - \overline{\mathbf{f}}_k||^2_2,
\end{eqnarray}
which is therefore a differentiable function of grid's vertex positions.
Our variance-based loss function aims to minimize the sum of variances across all grid cells:
\begin{eqnarray}
L_\text{var}(\mathbf{v}) = \sum\nolimits_{k=1}^K \tilde{V}_k(\mathbf{v})
\end{eqnarray}

\paragraph{Differentiable Reconstruction:}
Inspired by SSN~\cite{ssn}, we further differentiably reconstruct an image using the deformed grid by taking into account the probability of assignments: $P_{i\to k}(\mathbf{v})$. Intuitively, we represent each cell using its mean feature $\overline{\mathbf{f}}_k$, and ``paste" it back into the image plane according to the positions of the cell's deformed vertices. Specifically, we reconstruct each pixel in an image by softly gathering information from each grid cell using $P_{i\to k}(\mathbf{v})$:
\begin{eqnarray}
\hat{\mathbf{f}}_{i}(\mathbf{v})  &= \sum\nolimits_{k=1}^K P_{i\to k}(\mathbf{v})  \cdot \overline{\mathbf{f}}_k, \label{eq:paste_back} 
\end{eqnarray}
The reconstruction loss is the distance between the reconstructed pixel feature and original pixel feature:
\begin{eqnarray}
L_\text{recons}(\mathbf{v}) &= \sum\nolimits_{i=1}^N ||\hat{\mathbf{f}}_{i}(\mathbf{v})   - \mathbf{f}_i||_1.
\end{eqnarray}
We experimentally found that L1 distance works better than L2.



\paragraph{Regularization:}
To regularize the shape of the grid and prevent self-intersections, we introduce two regularizers. We employ an \textbf{Area balancing} loss function that encourages the areas of the cells to be similar, and thus, avoids self-intersections by minimizing the variance of the areas:
\begin{eqnarray}
L_\text{area}(\mathbf{v})  = \sum\nolimits_{j=1}^K ||a_k(\mathbf{v})  - \overline{a}(\mathbf{v}) ||_2^2,
\end{eqnarray}
where $\overline{a}$ is the mean area and $a_k$ is the area of cell $C_k$. We also utilize \textbf{Laplacian regularization} following works on 3D mesh prediction~\cite{pixel2mesh,dibr}. In particular, this loss encourages the neighboring vertices to move along similar directions with respect to the center vertex:
\begin{eqnarray}
L_\text{lap}(\mathbf{v}) = \sum\nolimits_{i=1}^n||\Delta_i -\frac{1}{||\mathcal{N}(i)||}\sum\nolimits_{j \in\mathcal{N}(i)}\Delta_{j}||^2_2,
\end{eqnarray}
where $\Delta_i = [\Delta x_i, \Delta y_i]^T$ is the predicted offset of vertex $v_i$ and $\mathcal{N}(i)$ is the set of neighboring vertices of vertex $v_i$.

The final loss to train  our network $h$ is a weighted sum of all the above terms:
\begin{eqnarray}
L_\text{def} =  L_\text{var} + \lambda_\text{recons} L_\text{recons} + \lambda_\text{area} L_\text{area} + \lambda_\text{lap} L_\text{lap}, 
\label{eq:grid_loss}
\end{eqnarray}
where $\lambda_{\text{recons}}, \lambda_{\text{area}}, \lambda_{\text{lap}}$ are hyperparameters that balance different terms.


\section{Applications}
\label{sec:applications}
Our {\ourmodel} supports many computer vision tasks that are done on fixed image grids today. We discuss three possible use cases in this section. 
{\ourmodel} can be inserted as a plug-and-play module at several levels of processing. By inserting it at the input level we utilize {\ourmodel} as a learnable geometric downsampling layer to replace standard pooling methods. We showcase its effectiveness through an application to semantic segmentation in Section~\ref{sec:downsampling}. We further show an application to object mask annotation in Section~\ref{sec:annotation} where we propose a model that reasons on the boundary-aligned grid output by a deep {\ourmodel} to produce object polygons. Lastly, we showcase {\ourmodel} as a standalone module for unsupervised image partitioning in Section~\ref{sec:superpixels}. 


\begin{figure}[t!]
\centering
\includegraphics[width=\linewidth]{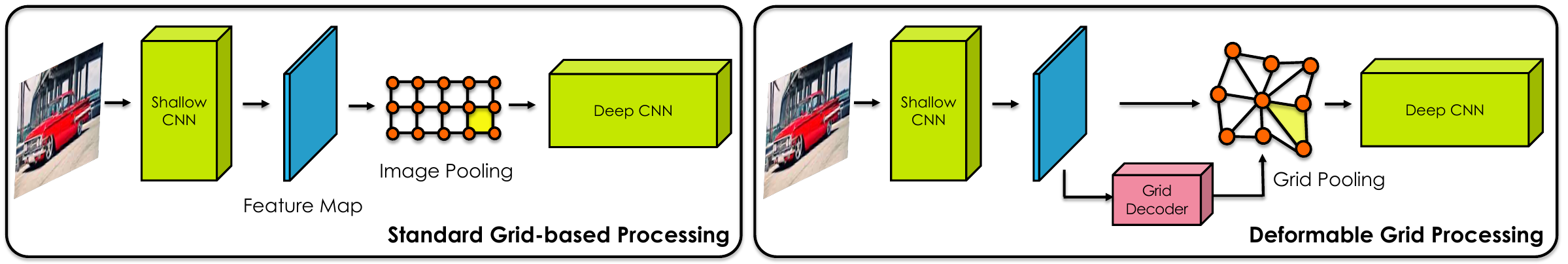}
\caption{ Feature pooling in fixed grid versus {\ourmodel}. {\ourmodel} can easily be used in existing deep CNNs and perform learnable downsampling.}
\label{fig:downsampling}
\end{figure}

\begin{figure}[t!]
\centering
\begin{minipage}{0.6\linewidth}
\includegraphics[width=\linewidth]{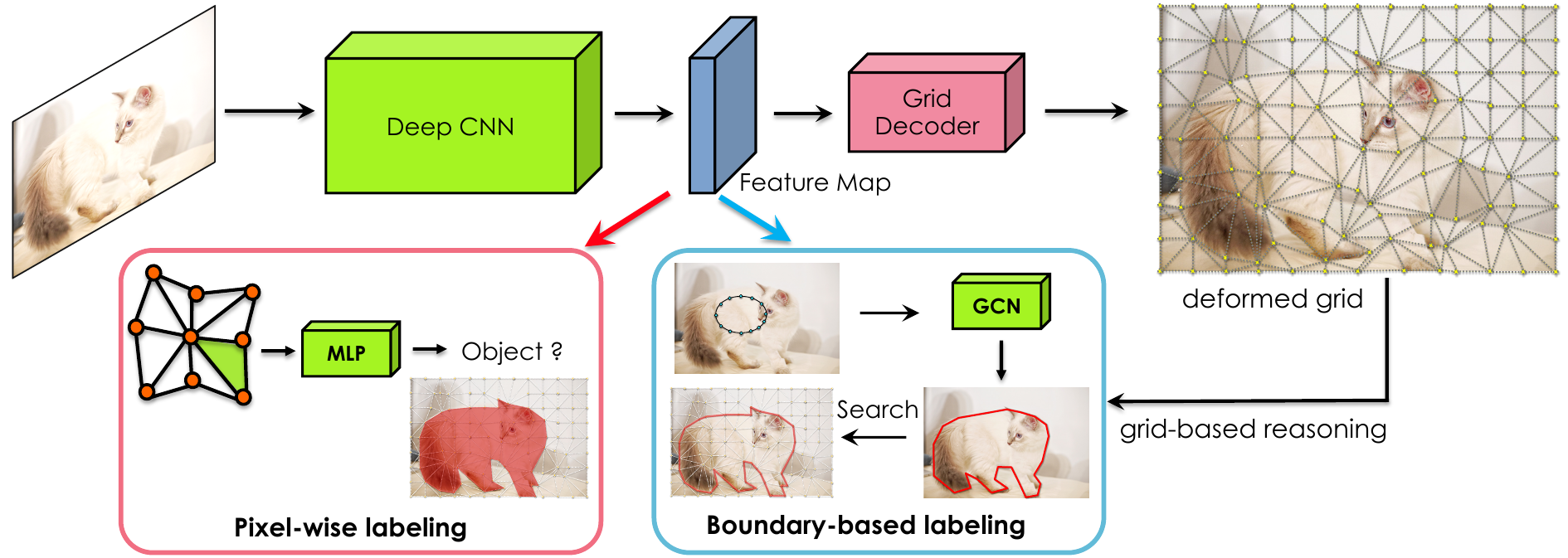}
\end{minipage}
\hfill
\begin{minipage}{0.32\linewidth}
\caption{ Object mask annotation by reasoning on a {\ourmodel}'s boundary-aligned grid. We support both pixel-wise labeling and curve-based tracing. }
\label{fig:maskannot}
\end{minipage}
\end{figure}

\subsection{Learnable Geometric Downsampling}
\label{sec:downsampling}

Semantic segmentation of complex scenes typically requires high resolution images as input and thus produces high resolution feature maps which are computationally intensive. Existing deep CNNs often take downsampled images as input and use feature pooling and bottleneck structures to relieve the memory usage~\cite{resnet,densenet,pspnet}. We argue that downsampling the features with our {\ourmodel} can preserve finer geometric information.
Given an arbitrary deep CNN architecture, we propose to insert a {\ourmodel} using a shallow CNN encoder to predict a deformed grid.
The predicted boundary-preserving grid can be used for geometry-aware feature pooling. 
Specifically, to represent each cell we can apply mean or max pooling by averaging or selecting maximum feature values within each triangular cell. 
Due to the regular grid topology, these features can be directly passed to a standard CNN. 
Note that the grid pooling operation warps the original feature map from the image coordinates to grid coordinates.
Thus the final output (predicted semantic segmentation) is pasted back into the image plane by checking in which grid's cell the pixel lies. 
The full pipeline is end-to-end differentiable. We can jointly train the model in a multi-task manner with a cross-entropy loss for the semantic segmentation branch and grid deformation loss in Eq.~\ref{eq:grid_loss}. 
The {\ourmodel} module is lightweight and thus bears minimal computational overhead. The architecture is illustrated in Fig~\ref{fig:downsampling}.

\subsection{Object Mask Annotation}
\label{sec:annotation}
Object mask annotation is the problem of outlining a 
foreground object given a user-provided bounding box~\cite{polyrnn,polyrnnpp,curvegcn,dextr,delse}.
Two dominant approaches have been proposed to tackle this task. The first approach utilizes a deep neural network to predict a pixel-wise mask~\cite{dextr,delse,polytransform}. The second approach tries to outline the boundary with a polygon/spline~\cite{huang2018deepprimitive,gao2019deepspline,polyrnn,polyrnnpp,curvegcn}. Our {\ourmodel} supports both approaches, and improves upon them via a polygonal grid-based reasoning (Fig.~\ref{fig:maskannot}).  


\textbf{Boundary-based segmentation:}
We formulate the boundary-based segmentation as a minimal energy path searching problem. We search for a closed path along the grid's edges that has minimal Distance Transform energy\footnote{The DT energy for each pixel is its distance to the nearest boundary.}: 
\begin{eqnarray}
Q = \argmin_{Q\in\mathcal{Q}} \sum\nolimits_{i=1}^{M} DT(v_{Q_i}, v_{Q_{(i+1)\%M}}),
\end{eqnarray}
where $\mathcal{Q}$ denotes the set of all possible paths on our grid, and $M$ is the length of path $Q$. We first predict a distance transform energy map for an object using a deep network trained with the L2 loss. We then compute the energy in each grid vertex via bilinear sampling. We obtain the energy for each grid edge by averaging the energy values for the points along the line defined by two vertices. Note that directly searching on the grid may result in many local minima. We employ Curve-GCN~\cite{curvegcn} to predict 40 seed points and snap each of these points to the grid vertex that has the minimal energy among its top-$k$ closest vertices. Then for each neighboring seed points pair, we use Dijkstra algorithm to find the minimal energy path between them. We provide algorithm details in the appendix. Our approach improves over Curve-GCN in two aspects: 1) it better aligns with image boundaries as it explicitly reasons on our boundary-aligned grid, 2) since we search for a minimal energy path between neighboring points output by Curve-GCN, our approach can handle objects with more complex boundaries that cannot be represented with only 40 points.  

\textbf{Pixel-wise segmentation:} Rather than producing a pixel-wise mask, we instead predict the class label for each grid cell. Specifically, we first use a deep neural network to obtain a feature map from the image. Then, for every grid cell, we average pool the feature of all pixels that are inside the cell, and use a MLP network to predict the class label for each cell. The model is trained with the cross-entropy loss. As the grid boundary aligns well with the object boundary, pooling the feature inside the grid is more efficient and effective for learning.

\subsection{Unsupervised Image Partitioning}
\label{sec:superpixels}
We can already view our deformed triangular cells as ``superpixels", trained with unsupervised loss functions. 
We can go further and cluster cells by using the affinity between them. In particular, we view the deformed grid as an undirected weighted graph where each grid cell is a node and an edge connects two nodes if they share an edge in the grid. The weight for each edge is the affinity between two cells, which can be calculated using RGB values of pixels inside the cells. 

Different clustering techniques can be used and exploring all is out of scope for this paper. To show the effectiveness of {\ourmodel} as an unsupervised image partitioning method, we here utilize simple greedy agglomerative clustering. We average the affinity to represent a new node after merging. 
Clustering stops when we reach the desired number of superpixels or the affinity is lower than a threshold. Note that our superpixels are, by design, polygons. Note that supervised loss functions are naturally supported in our framework, however we do not explore them in this paper.  

\section{Experiments}
We extensively evaluate {\ourmodel} on downstream tasks. We first show application to learnable downsampling for semantic segmentation.  We then evaluate on the object annotation task with boundary-based and pixel-wise methods. We finally show the  effectiveness of {\ourmodel} for unsupervised image partitioning.

\subsection{Learnable Geometric Downsampling}
\label{sec:exp_downsample}
To verify the effectiveness of our {\ourmodel} as an effective downsampling method, we compare it with (fixed) image grid feature pooling methods as baselines, namely max/average pooling and stride convolution, on the Cityscapes~\cite{cityscapes} semantic segmentation benchmark.  The baseline methods perform max/average pooling, or stride convolution on the shallow feature map, while our grid pooling methods apply the max/average pooling on deformed triangle cells. We compare our grid pooling with baselines when the height and width of the feature map is downsampled to 1/4, 1/8, 1/16 and 1/32 of the original image size. We use a modified ResNet50~\cite{resnet} which is more lightweight than SOTA models~\cite{Takikawa2019GatedSCNNGS}.

\paragraph{\textbf{Evaluation Metrics:}} Following~\cite{delse,polytransform,curvegcn}, we evaluate the performance using mean Intersection-over-Union (mIoU), and the boundary F score, with 4 and 16 pixels threshold on the full image. All metrics are averaged across all classes.



\begin{table*}[t!]
\begin{center}
\begin{adjustbox}{width=\textwidth}
\begin{tabular}{|l|c|c|c|c|c|c|c|c|c|c|c|c|}
\hline
Downsampling Ratio & \multicolumn{3}{c|}{1/4} & \multicolumn{3}{c|}{1/8} & \multicolumn{3}{c|}{1/16} & \multicolumn{3}{c|}{1/32}  \\
\hline
Metric & mIoU & F (4px) & F (16px) & mIoU & F (4px) & F (16px) & mIoU & F (4px) & F (16px) & mIoU & F (4px) & F (16px) \\
\hline\hline
Strided Convolution & 65.76 & 60.59 & 73.97 & 59.18 & 53.80 & 70.14 & 51.02 & 47.33 & 60.41 & 41.03 & 44.12 & 51.05 \\
Max Pooling & 66.32 & 60.92 & 74.18 & 59.83 & 54.22 & 70.71 & 52.93 & 49.00 & 63.59 & 42.44 & 45.23 & 52.47 \\
Average Pooling & 66.53 & 61.09 & 74.39 & 59.45 & 0.5361 & 70.67 & 52.01 & 47.58 & 61.33 & 43.54 & 45.37 & 53.39 \\
\hline
Grid Max Pooling & 67.87 & 64.37 & 75.10 & 64.75 & 61.02 & 72.95 & 55.87 & \textbf{53.98} & \textbf{66.62} & 47.20 & \textbf{48.74} & \textbf{60.73} \\
Grid Average Pooling & \textbf{67.91} & \textbf{64.99} & \textbf{75.43} & \textbf{65.36} & \textbf{61.12} & \textbf{73.03} & \textbf{56.94} & 53.77 & 66.38 & \textbf{48.30} & 48.67 & 60.62 \\
\hline
\end{tabular}
\end{adjustbox}
\caption{ {\bf Learnable downsampling} on Cityscapes Semantic Segm benchmark.}
\label{tbl:downsample}
\end{center}
\end{table*}
\begin{figure*}[t!]
\centering
\addtolength{\tabcolsep}{-4.0pt}
\begin{tabular}{cccc}
\includegraphics[height=1.52cm,trim=150 80 150 80,clip]{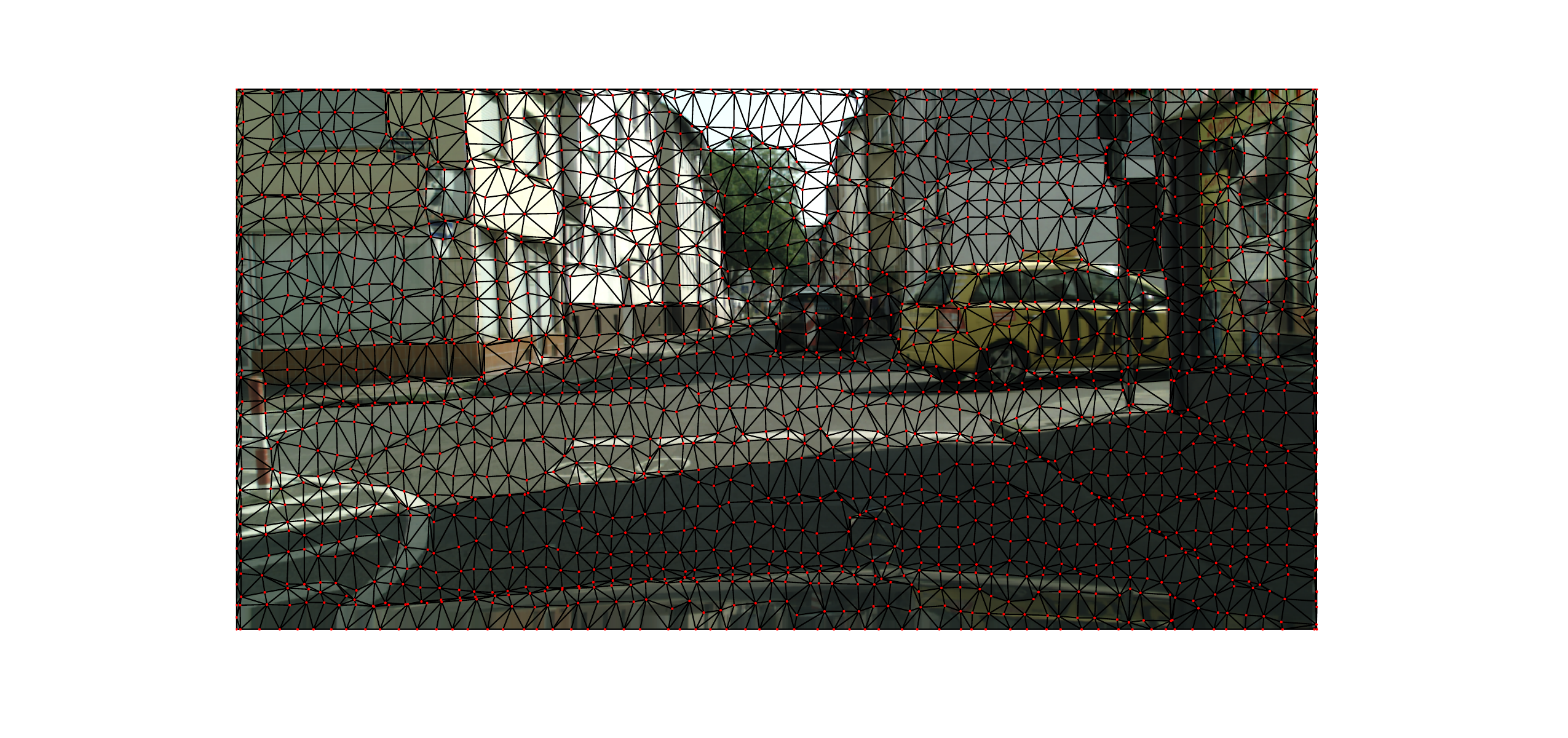} &
\includegraphics[height=1.52cm,trim=0 0 0 0,clip]{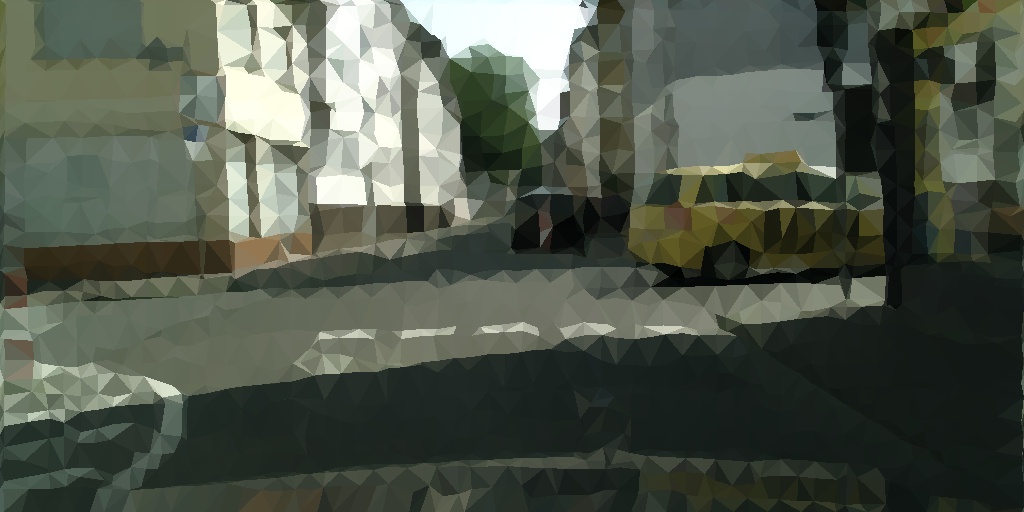}&
\includegraphics[height=1.52cm,trim=150 80 150 80,clip]{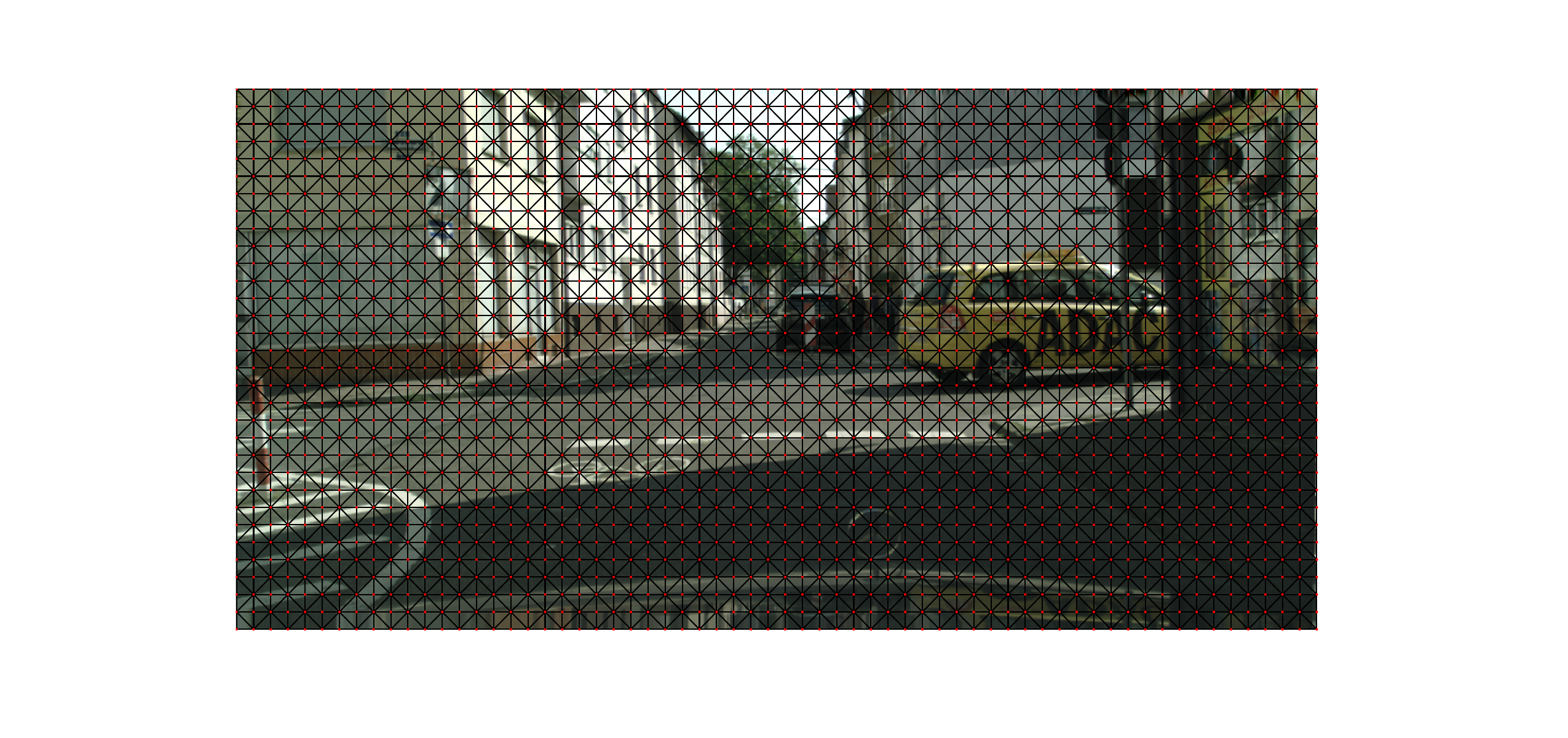} &
\includegraphics[height=1.52cm,trim=0 0 0 0,clip]{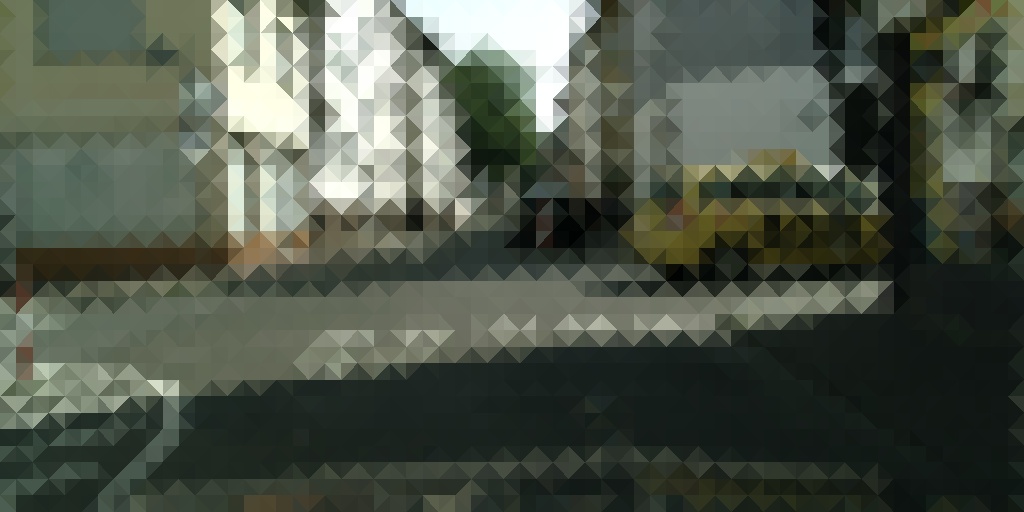}\\
[-1mm]
\includegraphics[height=1.52cm,trim=150 80 150 80,clip]{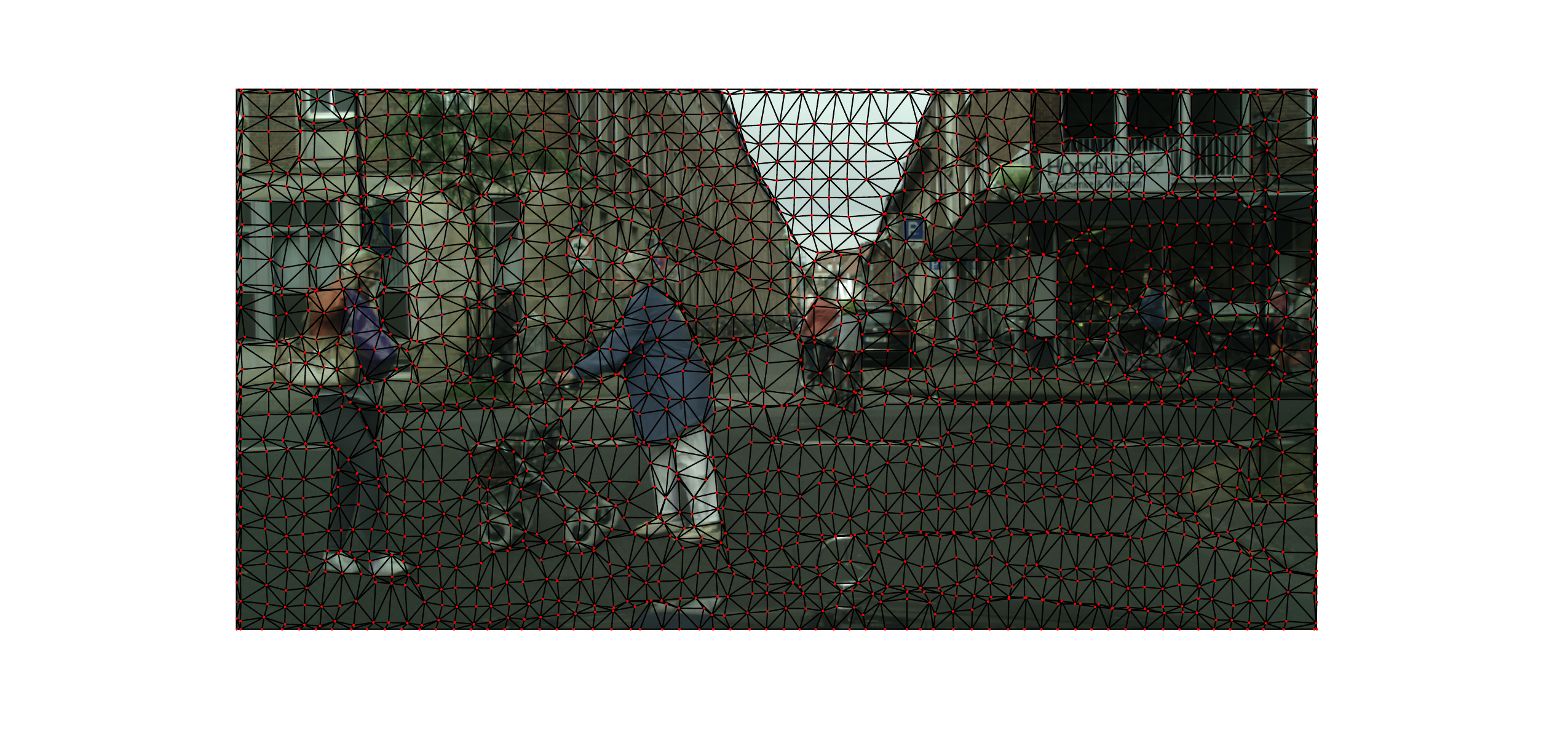} &
\includegraphics[height=1.52cm,trim=0 0 0 0,clip]{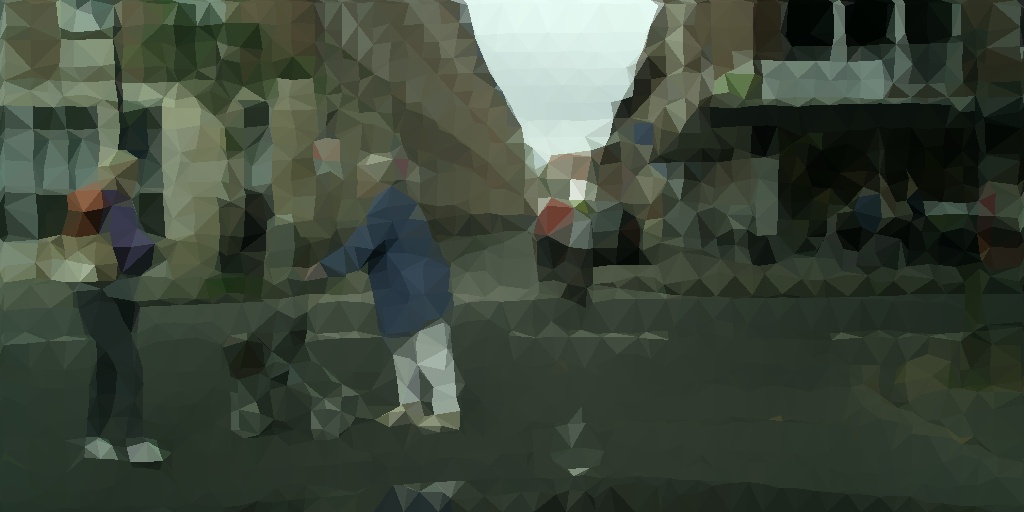} &
\includegraphics[height=1.52cm,trim=150 80 150 80,clip]{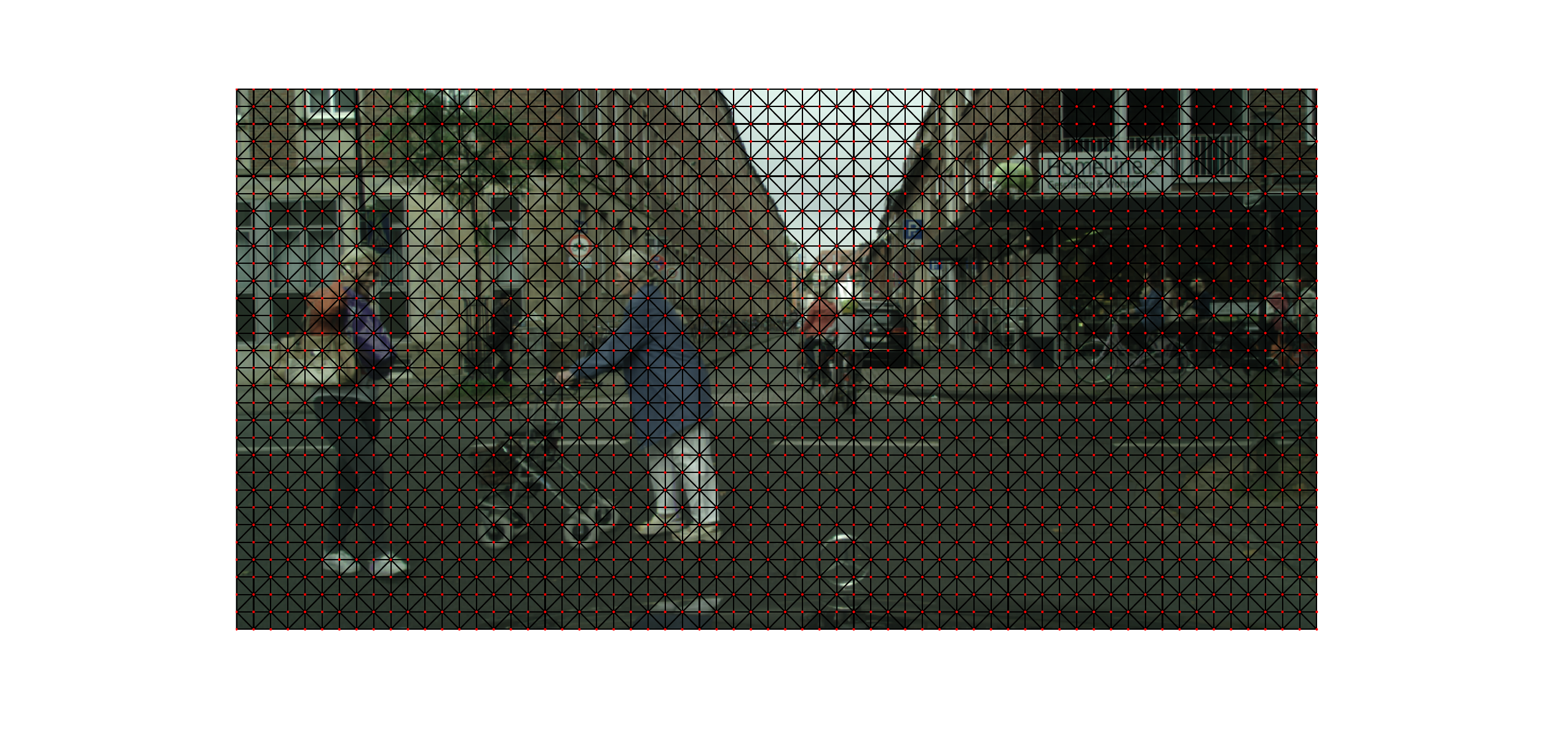} &
\includegraphics[height=1.52cm,trim=0 0 0 0,clip]{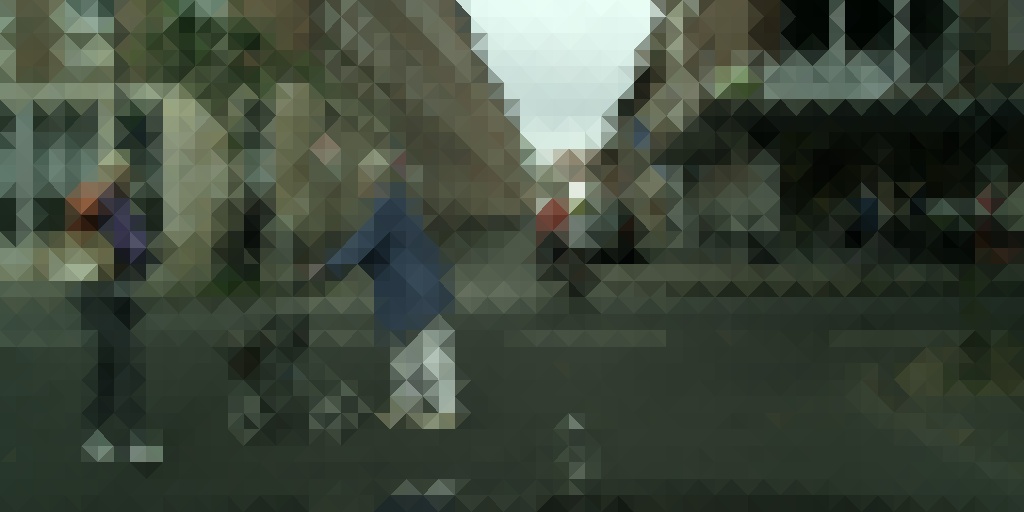} 
\end{tabular}
\caption{{\bf Illustration of learnable downsampling:} We show {\ourmodel} and its reconstructed image (left), comparing it to {\fixedgrid} (right). [Please zoom in]}
\label{fig:def_grid_city_full_image_seg}
\end{figure*}

\paragraph{\textbf{Results:}}
Performances (mIoU and boundary F scores) are reported in Table~\ref{tbl:downsample}. 
Our {\ourmodel} pooling methods consistently outperform the baselines, especially on the boundary score. We benefit from the edge-aligned property of the {\ourmodel} coordinates. 
At  1/8 with 1/4 downsampling ratios, the baseline performance drops significantly due to missing the tiny instances, while our {\ourmodel} pooling methods cope with this issue more gracefully. We also outperform baselines when the downsampling ratio is small, showing an efficient usage of limited spatial capacity. We visualize qualitative results for the predicted grids in Fig.~\ref{fig:def_grid_city_full_image_seg}. Our {\ourmodel} better aligns with boundaries and thus what the downstream network ``sees" is more informative than the fixed uniform grid. 

\begin{table*}[t!]
    \begin{center}
    {   
        \begin{adjustbox}{width=\textwidth}
        \addtolength{\tabcolsep}{1pt}
        \begin{tabular}{|l|c|c|c|c|c|c|c|c|c||c|c|}
            \hline
             Model & Bicycle & Bus & Person & Train & Truck & Mcycle & Car & Rider & mIoU & F1 & F2 \\
            \hline
            \hline
             Curve-GCN~\cite{curvegcn}& 75.40& 86.02& 79.87&82.89 & 86.44& 75.69& 90.21& 76.61& 81.64&59.45 & 75.43\\
            \hline
            \hline
             Pixel-wise&74.95 &86.19 &80.35 & 81.10& 86.10& 75.82&89.78 & 77.14 & 81.43 & 60.25 & 74.49 \\
            \hline
            \hline
             {\fixedgrid} &75.10 &85.73 & 79.84& 84.35&86.02 &75.97 &89.76& 76.56 & 81.67 & 59.01 &74.77\\
            {\ourmodel} &\textbf{75.46} &\textbf{86.29} & \textbf{80.40}&\textbf{84.91}&\textbf{86.58} & \textbf{76.13}& \textbf{90.42}& \textbf{77.20}\textbf{}&\textbf{82.17} &\textbf{61.94} & \textbf{77.04} \\
            \hline
        \end{tabular}
        \end{adjustbox}
        \caption{ {\bf Boundary-based object annotation} on Cityscapes-Multicomp.}
        \label{tbl:defgrid-ins-seg}
    }
    \end{center}
    
\end{table*}

\begin{figure*}[t!]
\addtolength{\tabcolsep}{-2.5pt}
\centering
\begin{tabular}{ccccc}
\includegraphics[height=2.4cm,trim=100 100 50 90,clip]{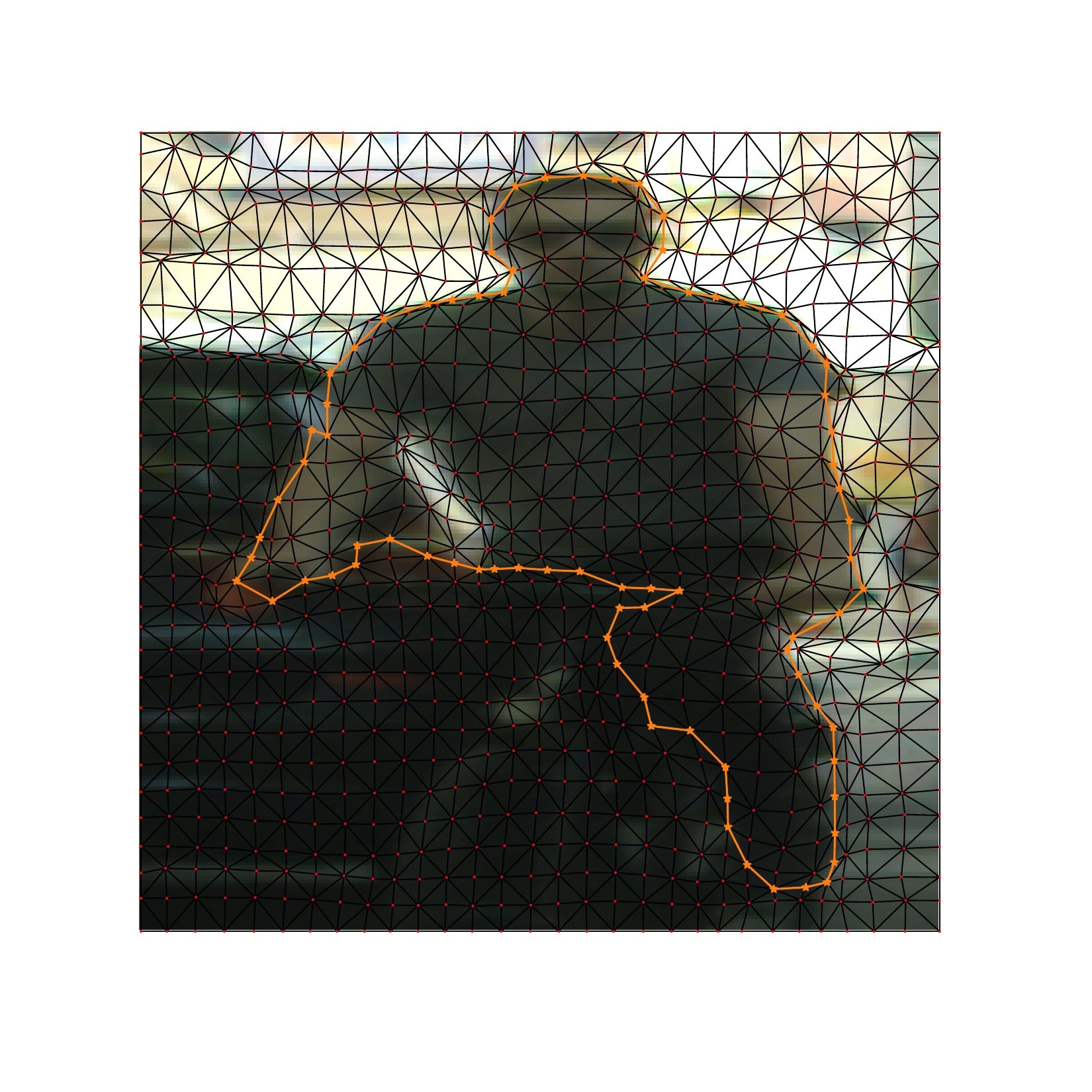} &
\includegraphics[height=2.4cm,trim=100 100 50 90,clip]{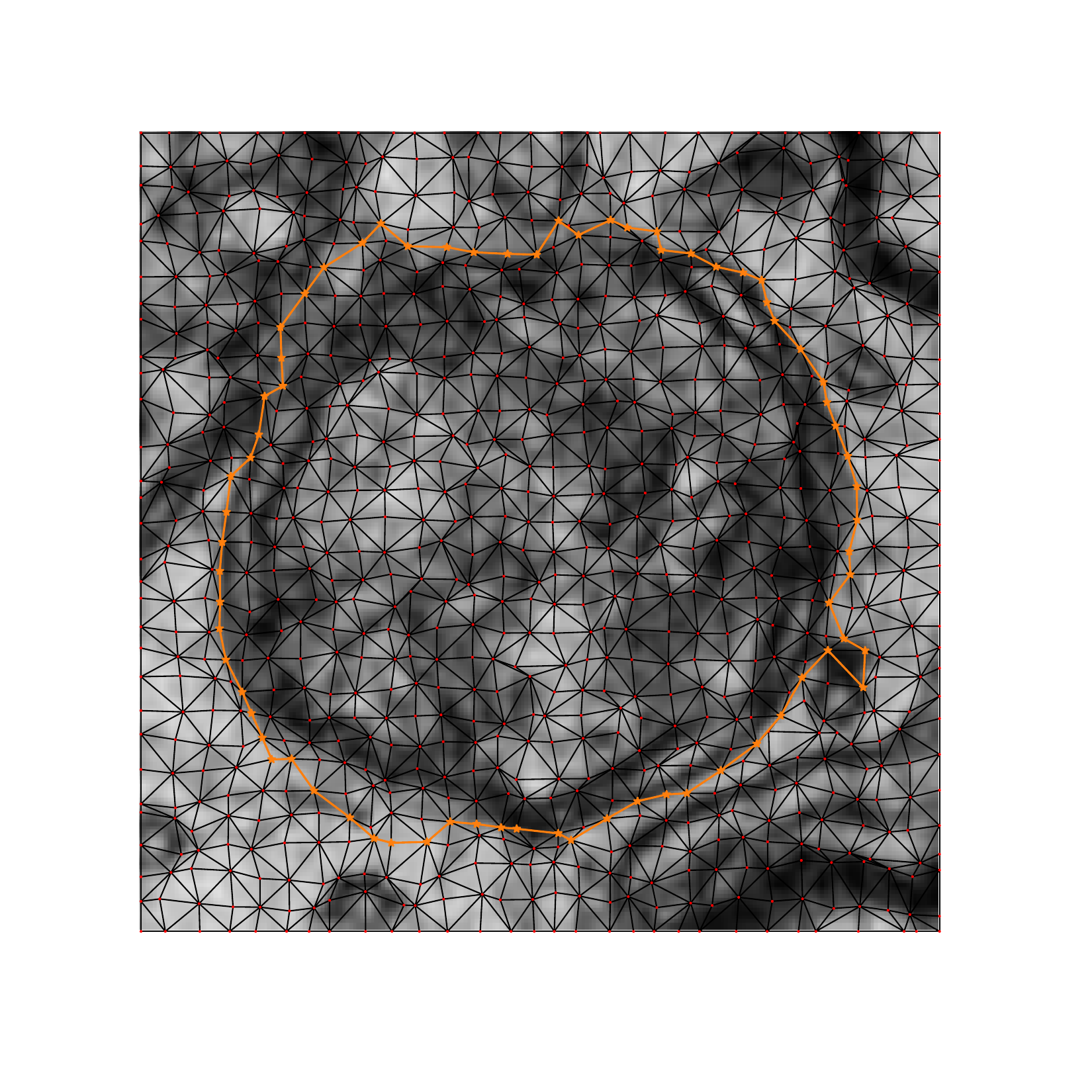} &
\includegraphics[height=2.4cm,trim=100 100 50 90,clip]{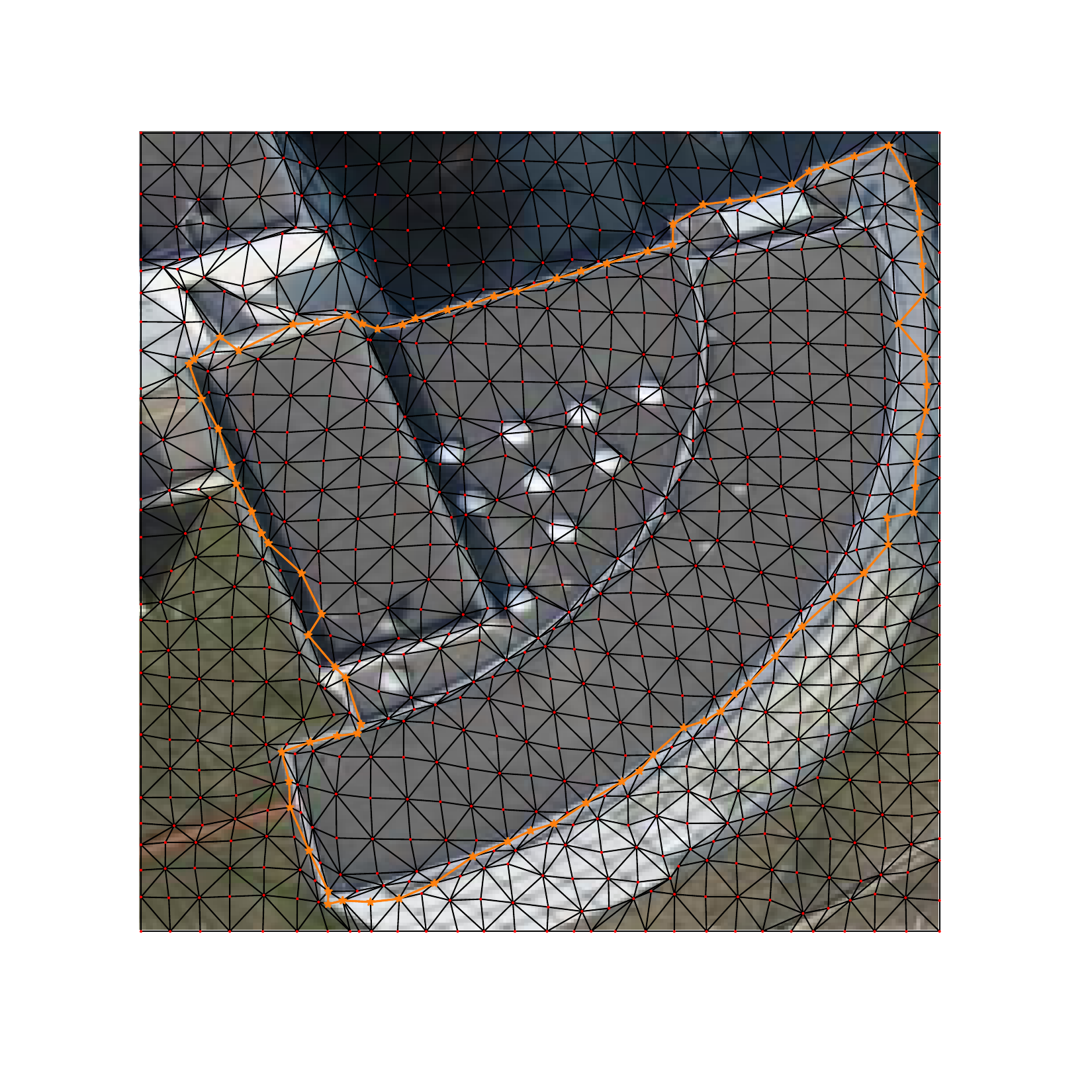} &
\includegraphics[height=2.4cm,trim=100 100 50 90,clip]{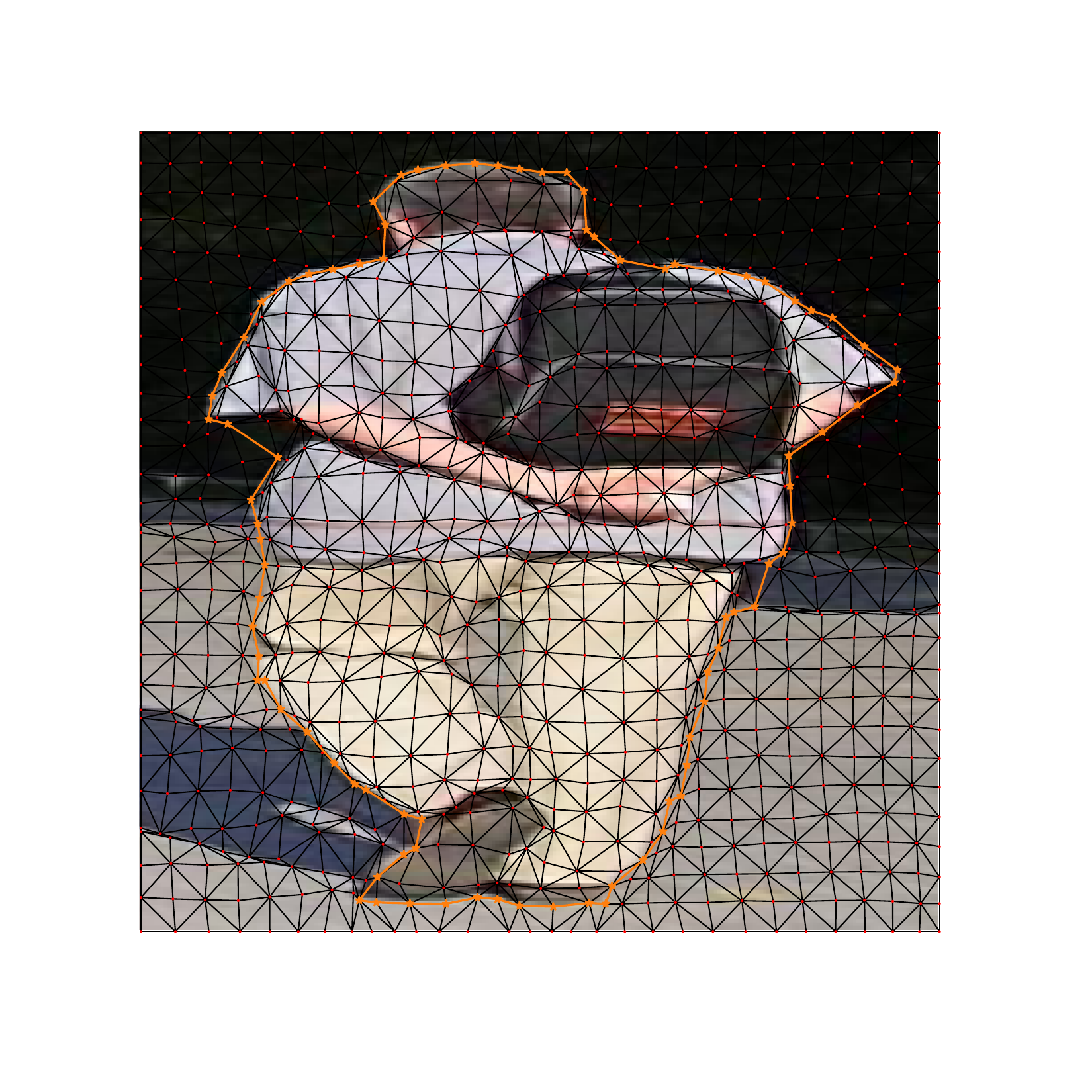} &
\includegraphics[height=2.4cm,trim=100 100 50 90,clip]{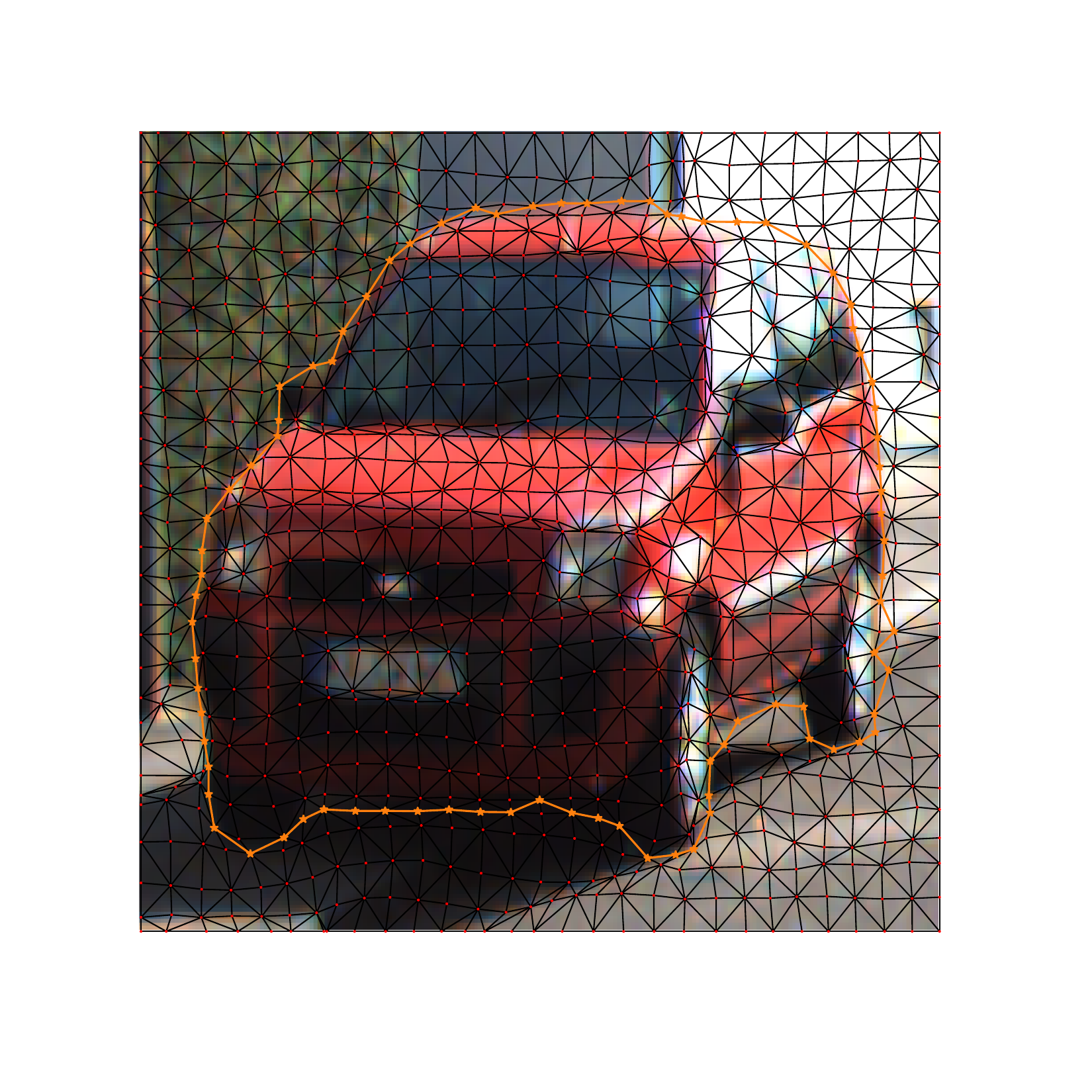}
\\[-1.7mm]
\includegraphics[height=2.4cm,trim=100 100 50 90,clip]{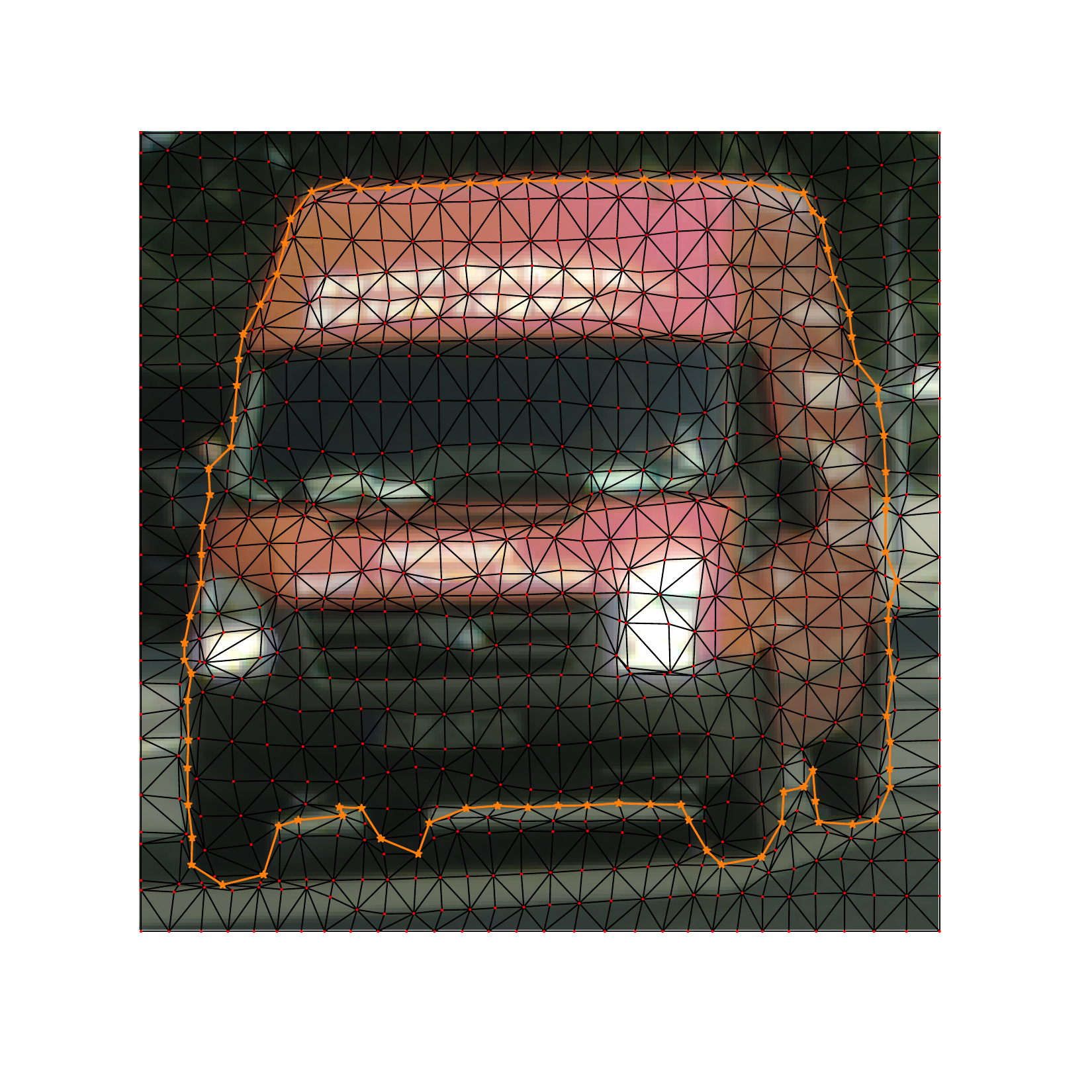} &
\includegraphics[height=2.4cm,trim=100 100 50 90,clip]{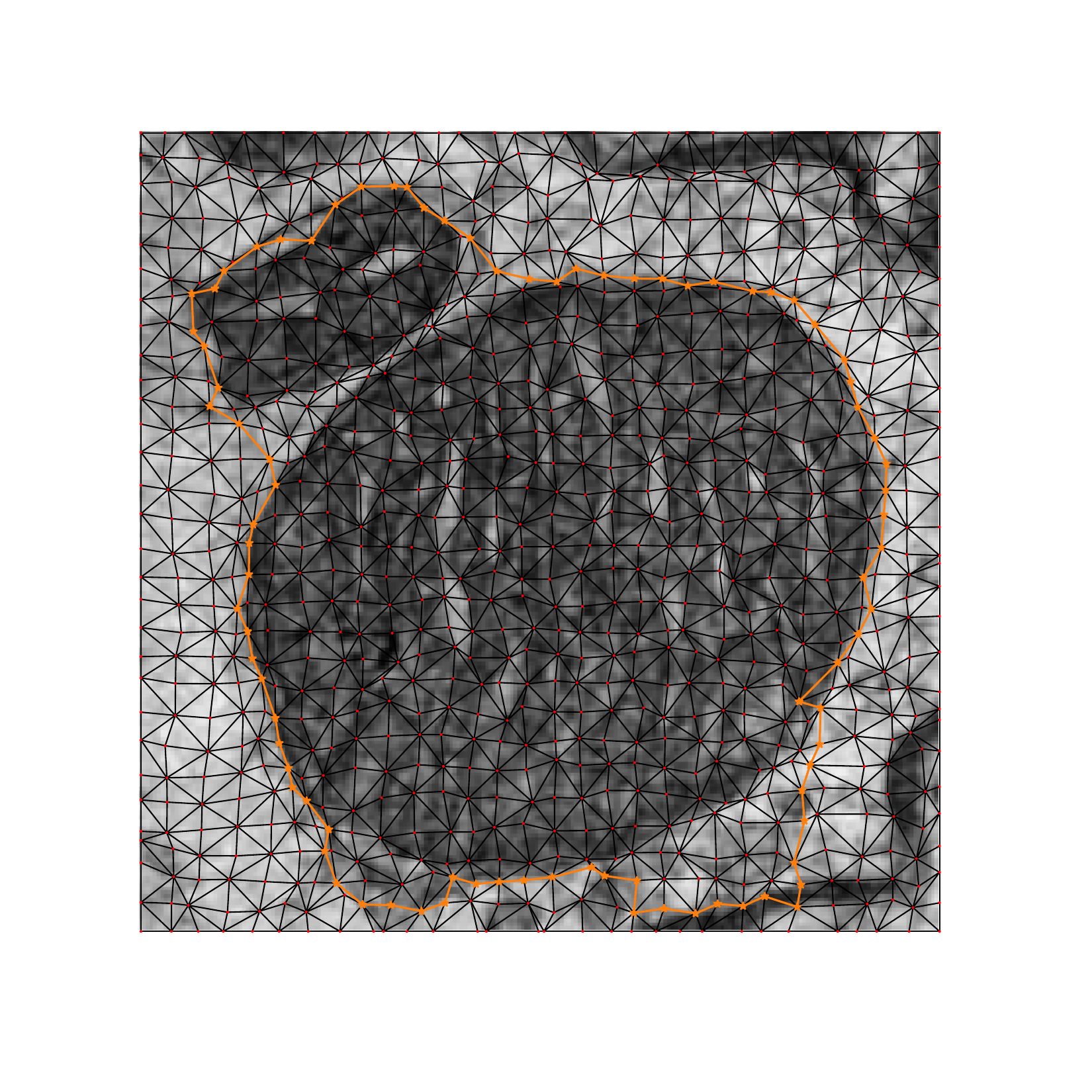} &
\includegraphics[height=2.4cm,trim=100 100 50 90,clip]{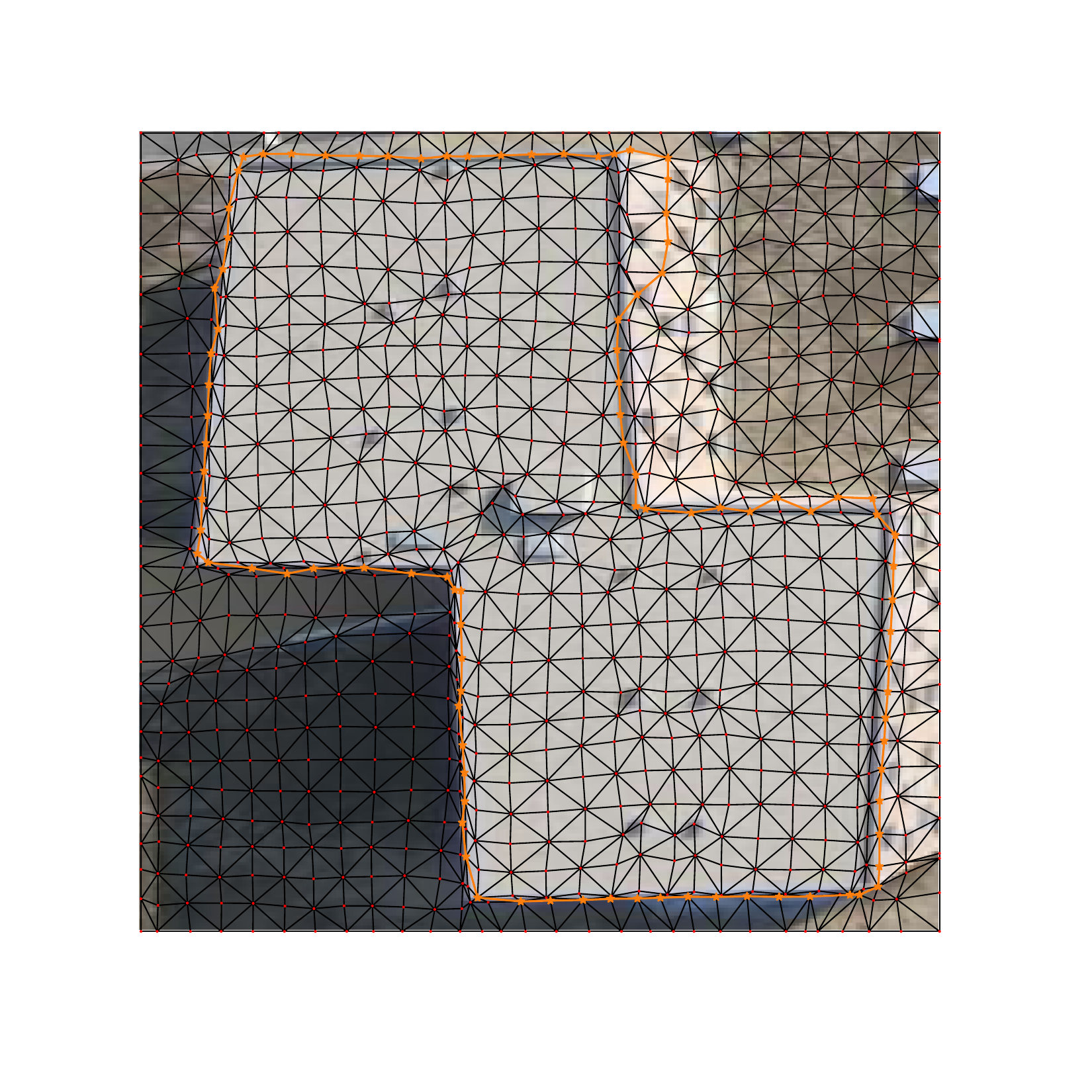} &
\includegraphics[height=2.4cm,trim=100 100 50 90,clip]{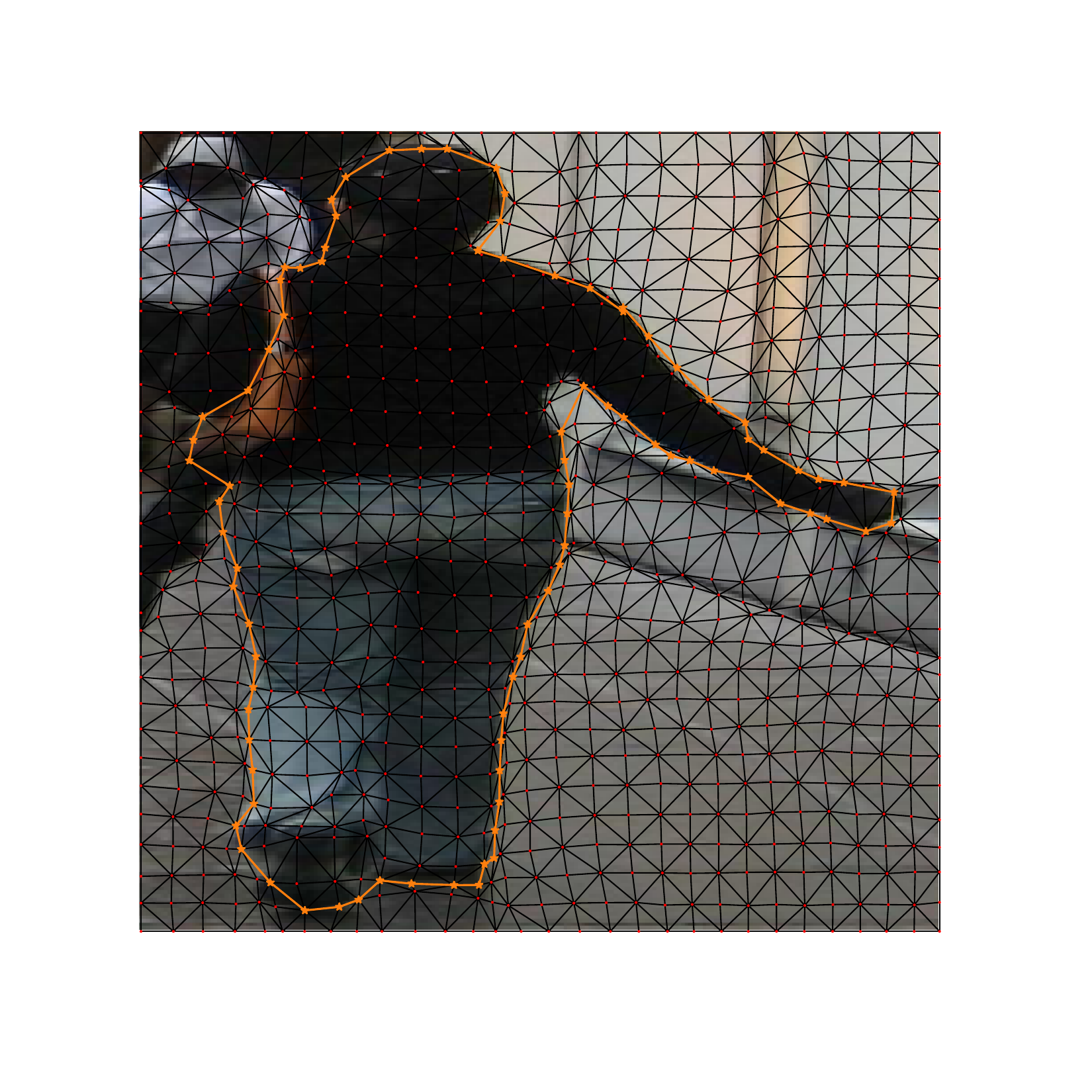} &
\includegraphics[height=2.4cm,trim=100 100 50 90,clip]{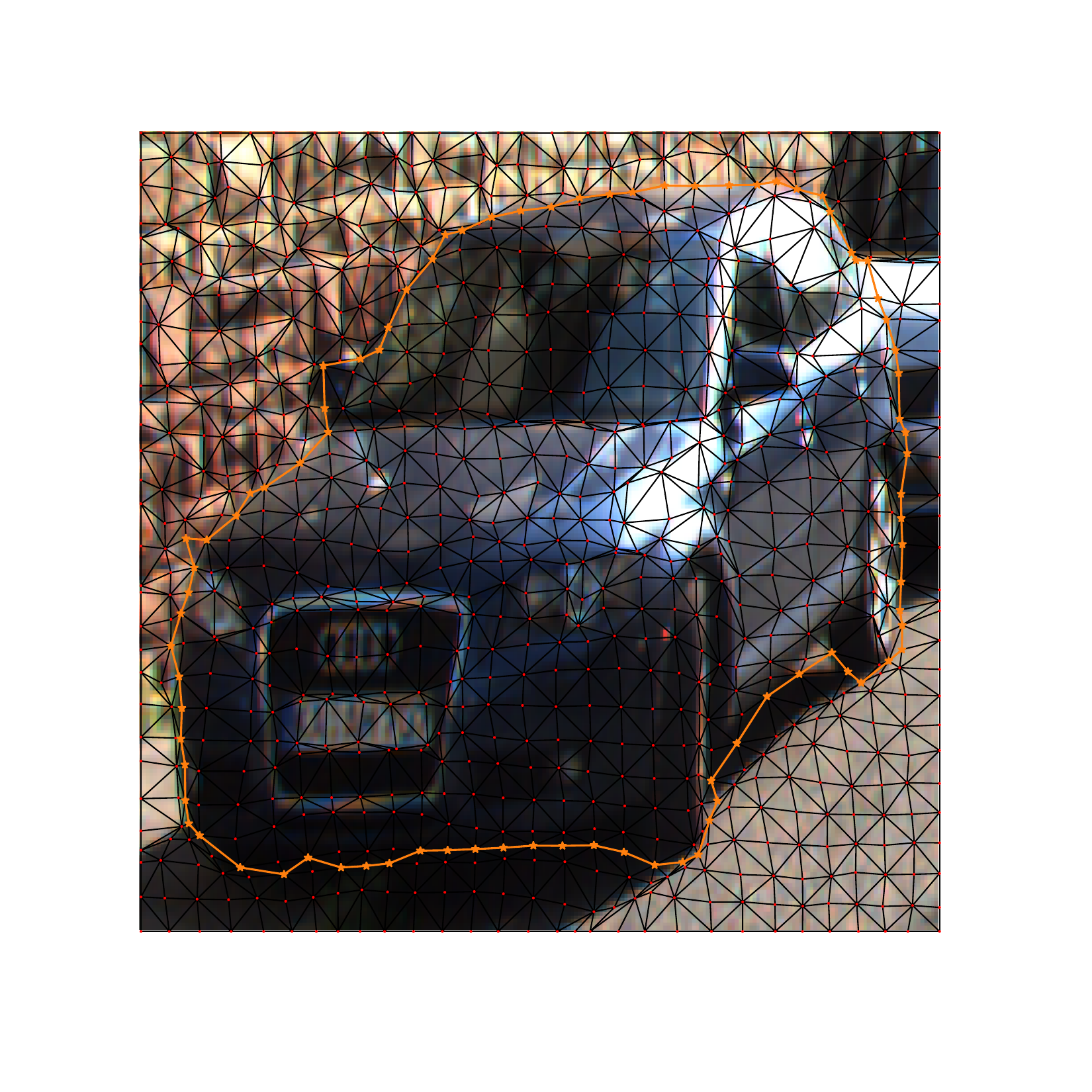} 
\end{tabular}
\caption{{\bf Deformed Grid:} We show examples of predicted grids both on in-domain (Cityscapes) and cross-domain images (Medical, Rooftop, ADE, KITTI). Orange line is the obtained minimal energy path along the grid's edges. 
}
\label{fig:def_grid_cross_domain}
\end{figure*}

\subsection{Object Annotation}
\label{sec:exp-boundary}
\paragraph{\textbf{Dataset:}} Following~\cite{polyrnn,polyrnnpp,curvegcn,delse,polytransform}, we train and test both of our instance segmentation models on the Cityscapes dataset~\cite{cityscapes}. We assume the bounding box of each object is provided by the annotator and the task is to trace the boundary of the object. We evaluate under two different settings, depending on the model. To compare with pixel-wise methods, we follow the setting proposed in DELSE~\cite{delse}, where an image is first cropped with a 10 pixel expansion around the ground truth bounding box, and resized to the size of 224$\times$224. This setting is referred to as Cityscapes-Stretch.
Our boundary-based annotation model builds on top of Curve-GCN which predicts a polygon that is topologically equivalent to a sphere. To compare with the baseline, we thus assume that the annotator creates a box around each connected component of the object individually.
We process each box in the same way as above, and evaluate performance on all component boxes. We refer to this setting as Cityscapes-Multicomp.

\begin{table*}[t!]
\begin{minipage}{0.62\linewidth}
    \begin{center}
    {
        \begin{scriptsize}
         \begin{adjustbox}{width=1\textwidth}
        \addtolength{\tabcolsep}{2pt}
        \begin{tabular}{|l|c|c|c|c|c|c|}
            \hline
            Method  & Metrics &  KITTI & ADE & Rooftop & Card.MR &ssTEM\\
            \hline
            \hline
            \multirow{3}{*}{Curve-GCN~\cite{curvegcn}}& mIoU & 87.43  & 76.71  & 81.11  & 86.18 & 68.97  \\
             & F1 &  64.90  &39.37 & 30.99 &  62.73 &  44.74 \\
              & F2 &  78.40  &  52.86& 45.08 & 78.22  &   59.85 \\
            \hline
            \multirow{3}{*}{Pixel-wise}& mIoU & 86.99 & 78.23 & 80.81 &  88.00& 69.37 \\
             & F1 &  62.73  & 42.88 & 28.44 & 62.63 & 46.07\\
              & F2 & 77.17 & 56.15 & 42.01& 77.94 & 59.77\\
             \hline
            \hline
            \multirow{3}{*}{\ourmodel}& mIoU &  \textbf{88.05}  & \textbf{78.54} &\textbf{83.10}& \textbf{89.01}  & \textbf{71.82} \\
             & F1 &  \textbf{66.70}  & \textbf{43.54} & \textbf{34.39} & \textbf{65.31}  & \textbf{50.31}  \\
             & F2 &  \textbf{80.23} & \textbf{57.20} & \textbf{49.79} & \textbf{81.92} & \textbf{65.14}  \\
            \hline
        \end{tabular}
        \end{adjustbox}
        \end{scriptsize}
       
            }
                \end{center}
        \end{minipage}
        \hfill
        \begin{minipage}{0.335\linewidth}
        \centering
         \caption{{\bf Cross-Domain Results} for boundary-based methods. {\ourmodel} significantly outperforms baselines, particularly evident in F-scores.}
        \label{tbl:def_grid_cross_domain}
        \end{minipage}
\end{table*}

\subsubsection{Boundary-based Object Annotation}
\paragraph{\textbf{Network Architecture:}} We use the image encoder from DELSE~\cite{delse}, 
and further add three branches to predict grid deformation, Curve-GCN points and Distance Transform energy map. For each branch, we first separately apply one 3$\times$3 conv filter to the feature map, followed by batch normalization~\cite{ioffe2015batch} and ReLU activation. For grid deformation, we extract the feature for each vertex with bilinear interpolation, and use a GCN to predict the offset for each vertex. For predicting spoints, we follow Curve-GCN~\cite{curvegcn}.
For the DT energy, we apply two 3$\times$3 conv filters with batch normalization and ReLU activation.

\paragraph{\textbf{Baselines:}} For Curve-GCN~\cite{curvegcn}, we compare with Spline-GCN, as it gives better performance than Polygon-GCN in the original paper~\cite{curvegcn}. We use the official  codebase 
but instead use our image encoder to get the feature map (with negligible performance gap) for a fair comparison. We also compare with pixel-wise methods, where we add two conv filters after the image encoder, which is similar to DELSE~\cite{delse} and DEXTR~\cite{dextr} but without extreme points.  We  further compare to our version of the model where the grid is fixed, referred to as {\fixedgrid}.
\begin{wrapfigure}[10]{r}{0.29\linewidth}
\centering
\includegraphics[width=1\linewidth, trim=10 50 10 80]{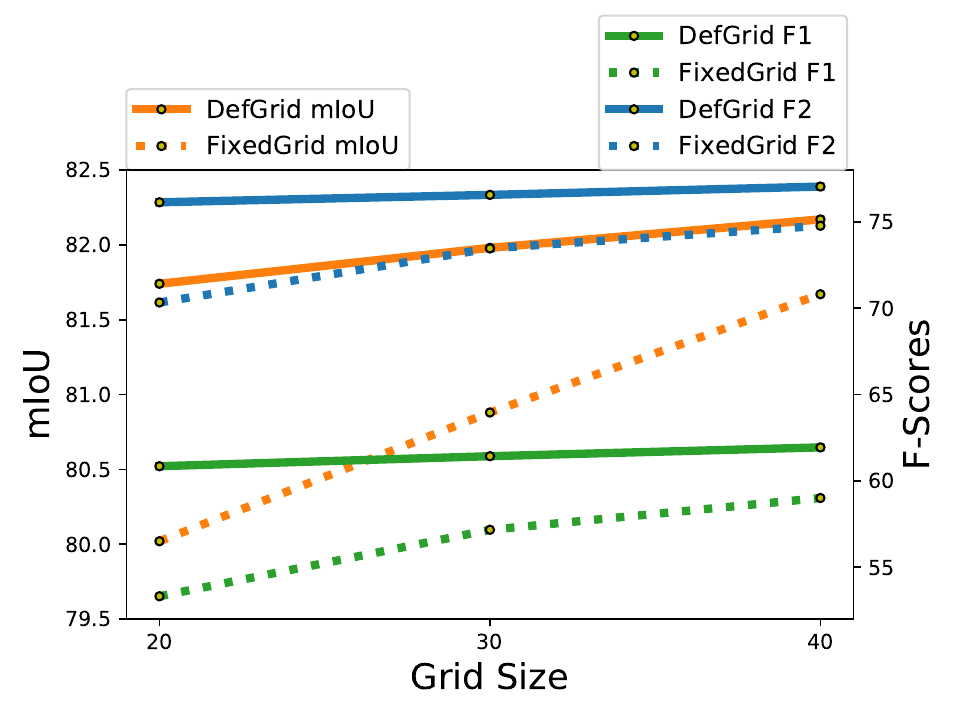}
\caption{ Object annotation for Cityscapes: {\fixedgrid} vs {\ourmodel}.}
\label{fig:obj_grid_sizs}
\end{wrapfigure}

\paragraph{\textbf{Results:}}
Table~\ref{tbl:defgrid-ins-seg} reports results. Our method outperforms baselines in all metrics. Performance gains are particularly significant in terms of F-scores, even when compared with pixel-wise methods. Since network details are the same across the methods, these results signify the importance of reasoning on the grid. Compared with {\fixedgrid} at different grid sizes in Fig.~\ref{fig:obj_grid_sizs}, {\ourmodel} achieves superior performance.  We attribute the performance gains due to better alignment with boundaries and more flexibility in tracing a longer contour.  We show qualitative results in Fig.~\ref{fig:def_grid_cross_domain} and~\ref{fig:def_grid_city_ins_seg}. 


\begin{table*}[t!]
\begin{center}
\begin{adjustbox}{width=\textwidth}
\begin{tabular}{|l|l|c|c|c|c|c|c|c|c|c|c|c|c|c|c|c|c|c|c|}
\hline
& & \multicolumn{3}{c|}{Round 0 (Automatic)}  & \multicolumn{3}{c|}{Round 1} & \multicolumn{3}{c|}{Round 2} & \multicolumn{3}{c|}{Round 3} & \multicolumn{3}{c|}{Round 4} & \multicolumn{3}{c|}{Round 5}  \\
\hline
Dataset &  Model & mIoU & F1 & F2 & mIoU & F1 & F2 & mIoU & F1 & F2 & mIoU & F1 & F2 & mIoU & F1 & F2 & mIoU & F1 & F2 \\
\hline\hline
\multirow{2}{*}{CityScapes~\cite{cityscapes}} & CurveGCN~\cite{curvegcn} & 80.74 & 58.01 & 73.67 &83.04 & 60.47 & 76.35 & 84.58 & 64.42 & 79.42 & 85.56 & 67.41 & 81.40 & 86.27 & 69.87 & 82.94 & 86.79 & 71.91 & 84.12 \\
 & \ourmodel &\textbf{81.98} & \textbf{61.43 }& \textbf{76.56 } & \textbf{84.48} &\textbf{ 66.59} & \textbf{80.96} & \textbf{85.73 }&\textbf{ 69.98} & \textbf{83.41} & \textbf{86.41 }& \textbf{72.05} & \textbf{84.71 }& \textbf{86.81} & \textbf{73.60} & \textbf{85.51} & \textbf{87.12} & \textbf{74.84 }& \textbf{86.20} \\
 \hline
 \hline
\multirow{2}{*}{ADE~\cite{ade20k}} & CurveGCN~\cite{curvegcn}  & 76.71 & 39.37 &  52.86 & 79.26 & 42.43 & 56.62 & 81.09 & 46.04 & 60.51 & 82.47 & 49.61 & 64.00 & 83.44 & 52.56 & 66.61 & 84.38 & 55.42 & 69.00 \\
 & \ourmodel & \textbf{78.54} & \textbf{43.54} &  \textbf{57.20} & \textbf{80.67 }& \textbf{48.57} &\textbf{ 62.76} & \textbf{82.27} & \textbf{52.33} & \textbf{66.45} & \textbf{83.43} & \textbf{55.60} & \textbf{69.37} &\textbf{ 84.23} &\textbf{ 58.22} &\textbf{ 71.79 }& \textbf{84.69 }& \textbf{60.03 }& \textbf{73.29} \\
 \hline
\multirow{2}{*}{KITTI~\cite{kitti}} & CurveGCN~\cite{curvegcn}& 87.43  &   64.90   & 78.40  & 89.41 & 69.20 & 82.84 & 90.50 & 73.11 & 85.84 & 91.22 & 75.94 & 87.73 & 91.67 & 78.16 & 89.20 & \textbf{92.00} & \textbf{79.90 }& 90.21 \\
 & \ourmodel & \textbf{88.05} &\textbf{ 66.70} & \textbf{80.23} & \textbf{89.94} & \textbf{70.80 }& \textbf{84.16} & \textbf{90.90} & \textbf{74.65} & \textbf{87.12} & \textbf{91.45} &\textbf{ 77.08} & \textbf{88.81} & \textbf{91.70} & \textbf{78.67} & \textbf{89.78} &91.87 & 79.72 &\textbf{ 90.33} \\
 \hline
\multirow{2}{*}{Rooftop~\cite{rooftop}} & CurveGCN~\cite{curvegcn} &  81.11 & 30.99 & 45.08   & 83.94 & 34.00 & 49.24 & 85.53 & 37.42 & 53.39 & 86.72 & 41.39 & 57.85 & 87.68 & 45.29 & 61.88 & 88.28 & 48.96 & 65.60 \\
 & \ourmodel & \textbf{83.10} & \textbf{34.39 }& \textbf{49.79} &\textbf{85.37} & \textbf{39.60 }& \textbf{56.04 }& \textbf{86.77 }& \textbf{43.74} & \textbf{60.73} & \textbf{87.53} & \textbf{47.30 }& \textbf{64.51} & \textbf{88.19} &\textbf{ 49.95} & \textbf{67.50} & \textbf{88.55} & \textbf{52.20} & \textbf{69.66} \\
 \hline
 \multirow{2}{*}{Card.MR~\cite{cardiacmr}} & CurveGCN~\cite{curvegcn}&86.18 &  62.73 & 78.22   & 89.22 & 68.26 & 84.91 & 90.42 & 73.10 & 89.16 & 91.25 & 76.94 & 91.94 & 92.00 & 80.67 & 94.45 & 92.45 & 83.45 & 95.99 \\
 & \ourmodel &\textbf{89.01} & \textbf{65.31} & \textbf{81.92}\textbf{} & \textbf{91.01} & \textbf{72.28} & \textbf{88.83} & \textbf{91.90} & \textbf{77.64} & \textbf{92.74 }& \textbf{92.55} & \textbf{81.03 }& \textbf{94.89} & \textbf{92.96 }& \textbf{83.24} &\textbf{ 95.98} & \textbf{93.31} & \textbf{85.47 }& \textbf{96.96} \\
 \hline
\multirow{2}{*}{ssTEM~\cite{sstem}} & CurveGCN~\cite{curvegcn} & 68.97 & 44.74 &  59.85& 71.66 & 47.81 & 63.21 & 74.75 & 52.78 & 68.82 & 76.57 & 56.88 & 72.99 & 78.06 & 60.67 & 76.50 & 79.22 & 63.84 & 79.17 \\
& \ourmodel  &\textbf{ 71.82} &\textbf{ 50.31} & \textbf{65.14} & \textbf{73.51} &\textbf{ 55.43} & \textbf{70.44} & \textbf{75.25} & \textbf{60.64} & \textbf{75.69} & \textbf{77.28} & \textbf{65.18} & \textbf{79.72} & \textbf{78.46} & \textbf{68.44} & \textbf{82.24} & \textbf{78.89} & \textbf{70.59} & \textbf{83.62} \\
\hline
\end{tabular}
\end{adjustbox}
\caption{ {\bf Interactive annotation results} on both in-domain (first two lines) and cross domain datasets. Our {\ourmodel} starts with a higher automatic performance and keeps this gap across (simulated) annotation rounds.
}
\label{tbl:interactive}
\end{center}
\end{table*}

\begin{table*}[t!]
    \begin{center}
    \begin{adjustbox}{width=\textwidth}
         \addtolength{\tabcolsep}{1.5pt}
        \begin{tabular}{|l|c|c|c|c|c|c|c|c|c|c|c|c|}
            \hline
            Model & Bicycle & Bus & Person & Train & Truck & Mcycle & Car & Rider & mIoU  & F1 & F2 \\
            \hline\hline
DELSE*~\cite{delse}&74.32&\textbf{88.85}&80.14&\textbf{80.35}&86.05&74.10&86.35&76.74&80.86&60.29&74.40\\
PolyTransform~\cite{polytransform}&74.22&88.78&80.73&77.91&86.45&\textbf{74.42}&86.82&\textbf{77.85}&80.90&62.33&76.55\\
OurBack + SLIC~\cite{slic} &73.88 & 85.47& 79.80& 77.97& 86.32& 72.62& 87.85&76.14 &80.01 &57.95 &72.17 \\
            \hline
            \hline
            \ourmodel  & \textbf{74.82} & 87.09 & \textbf{80.87} & \textbf{81.05} & \textbf{87.52} & 73.44 & \textbf{89.19} & 77.36 & \textbf{81.42} & \textbf{63.38} &\textbf{76.89}\\
            \hline
        \end{tabular}
      \end{adjustbox}
        \caption{{\bf Pixel-based methods} on Cityscapes-Stretch. Note that PolyTransform~\cite{polytransform} uses 512x512 resolution, while other methods use 224x224 resolution.}
        \label{tbl:defgrid-pixel-seg}
    \end{center}
\end{table*}
\paragraph{\textbf{Cross Domain Results:}}
Following~\cite{curvegcn}, we evaluate models' ability in generalizing to other datasets out of the box. Quantitative and qualitative results are  reported in Table~\ref{tbl:def_grid_cross_domain} and Fig.~\ref{fig:def_grid_cross_domain}, Fig~\ref{fig:def_grid_city_ins_seg}, respectively. We outperform all baselines on all datasets, in terms of all metrics, and the performance gains are particularly evident for the F-scores. Qualitatively, in all cross-domain examples the predicted grid's edges align well with real object boundaries (without any finetuning), demonstrating superior generalization capability for {\ourmodel}.

\paragraph{{\bf Interactive Instance Annotation:}} We follow Curve-GCN~\cite{curvegcn} and also report performance for the interactive setting in which an annotator corrects errors by moving the predicted polygon vertices. We follow the original setting but restrict our reasoning on the deformed grid. Results for different simulated rounds of annotation are reported in Table~\ref{tbl:interactive} with evident performance gains.

\begin{figure*}[t!]
\addtolength{\tabcolsep}{-1.0pt}
\begin{minipage}[t] {\textwidth}
\centering
\includegraphics[width=0.327\linewidth,trim=0 260 0 0,clip]{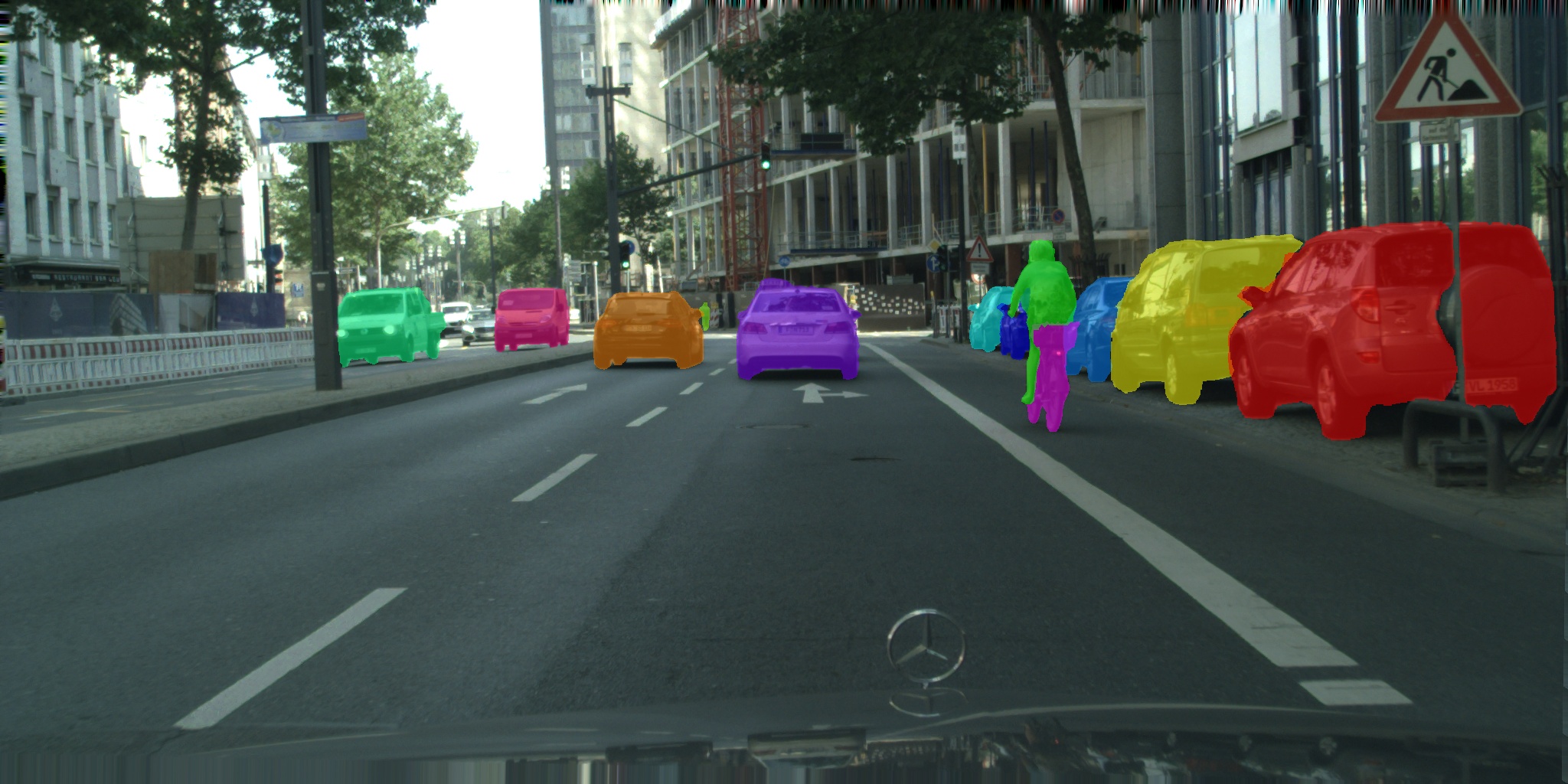} 
\includegraphics[width=0.327\linewidth,trim=0 260 0 0,clip]{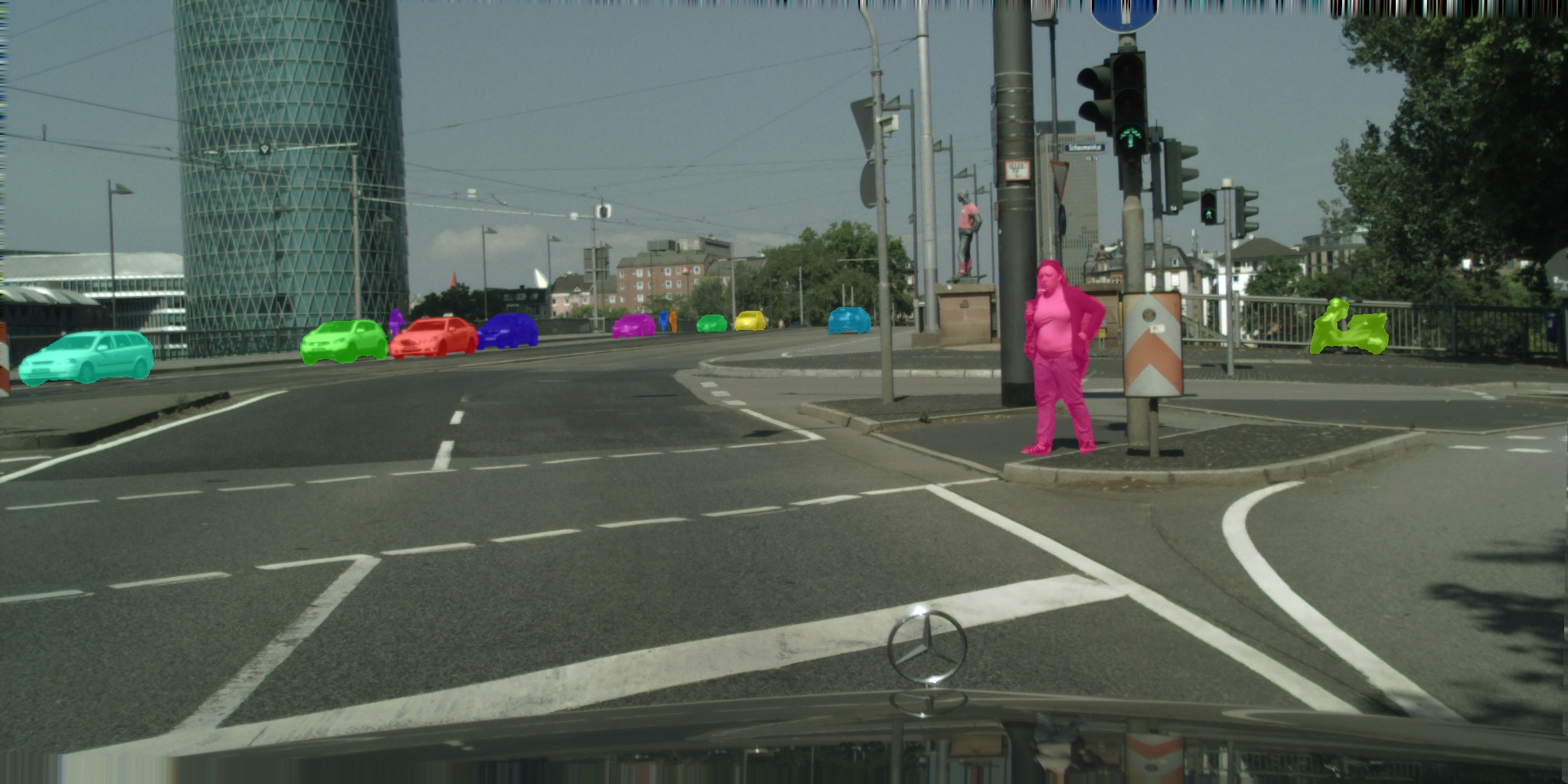}
\includegraphics[width=0.327\linewidth,trim=0 60 0 200,clip]{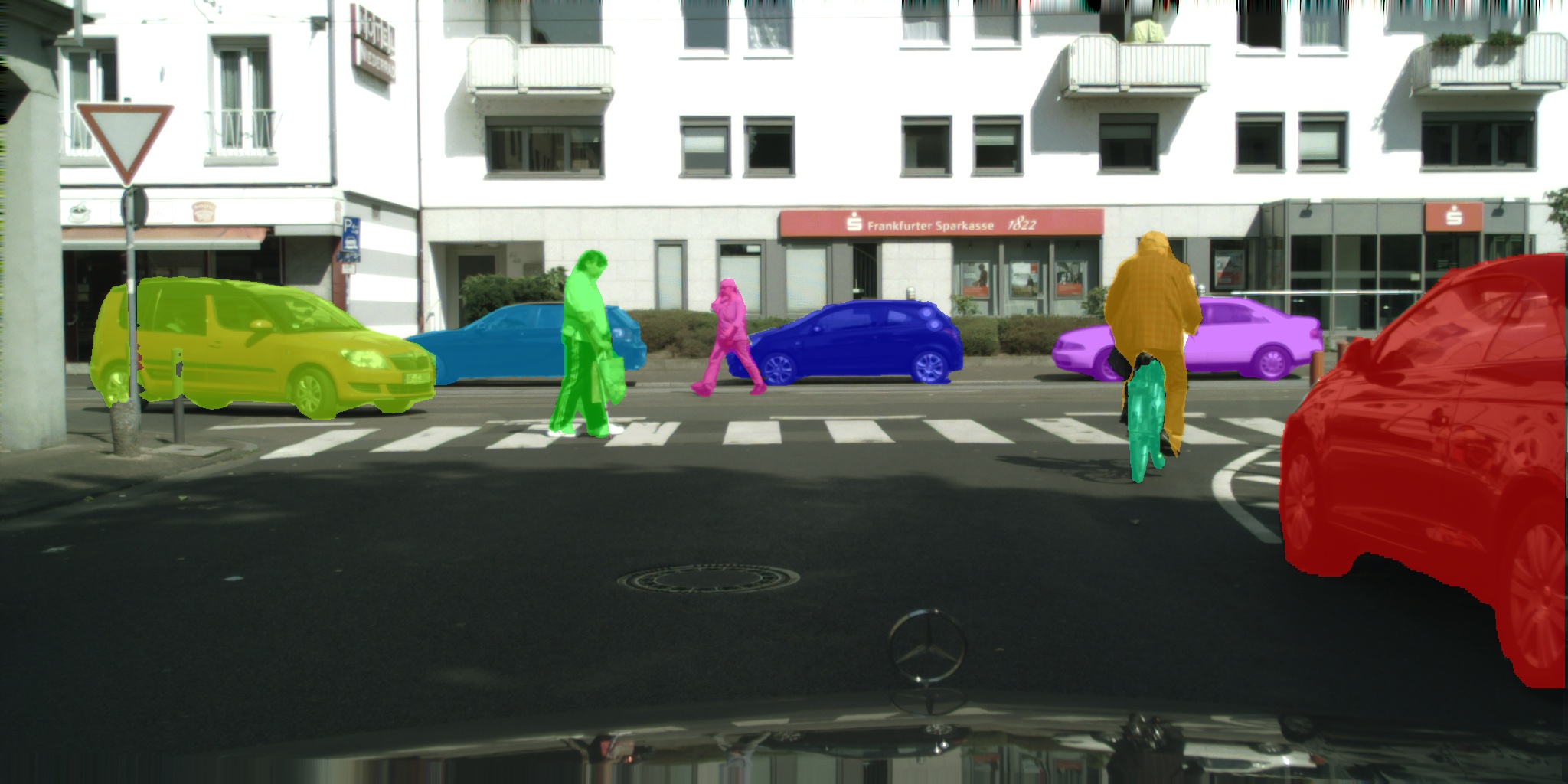} 
\end{minipage}
\begin{tabular}{cccc}
\includegraphics[height=1.75cm,trim=10 20 10 40,clip]{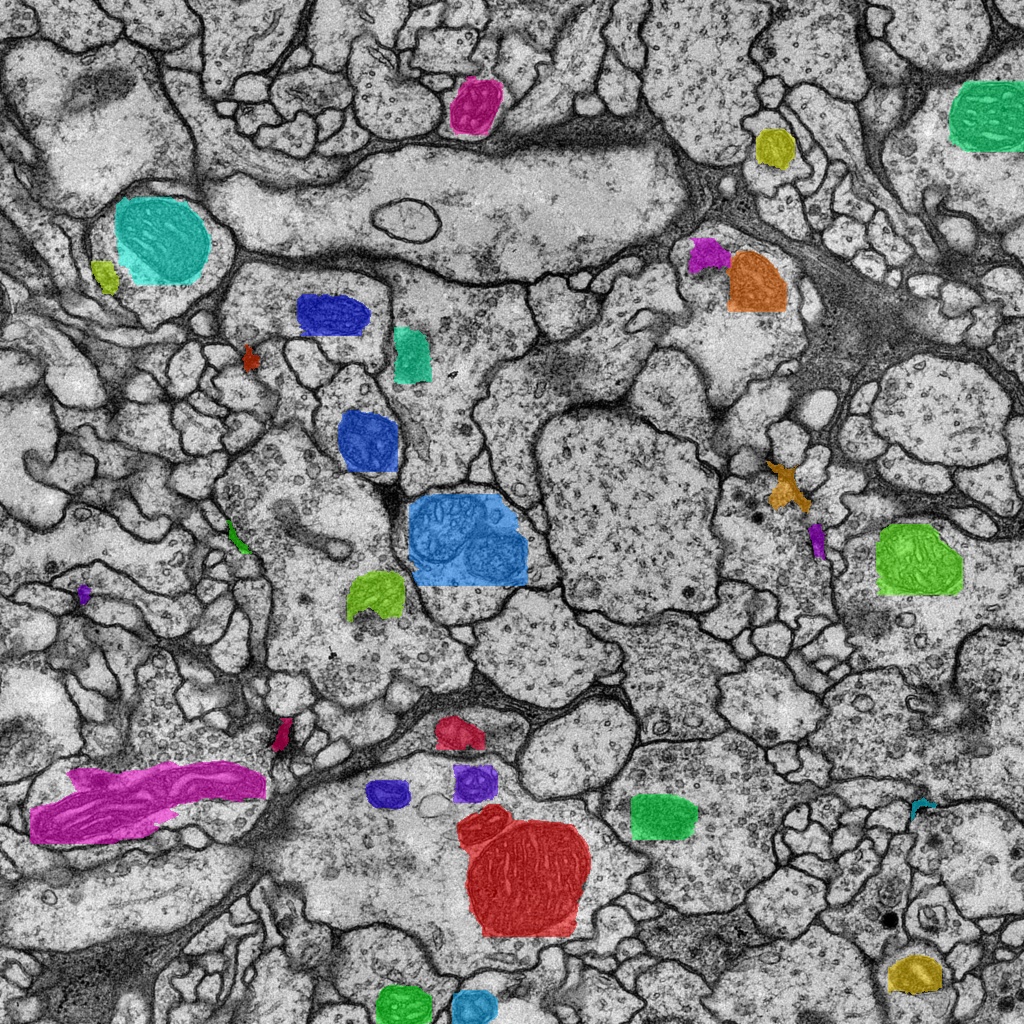} &
\includegraphics[height=1.75cm,trim=100 0 300 80,clip]{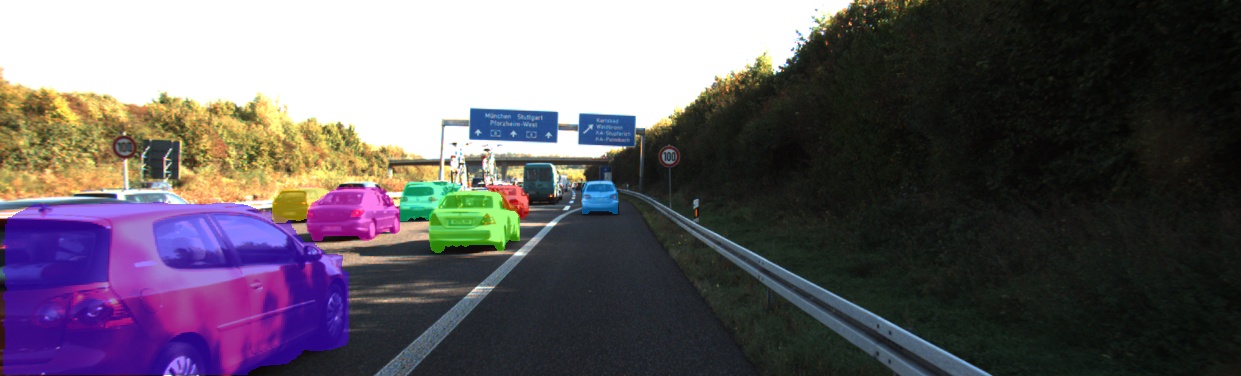} &
\includegraphics[height=1.75cm,trim=0 0 0 30,clip]{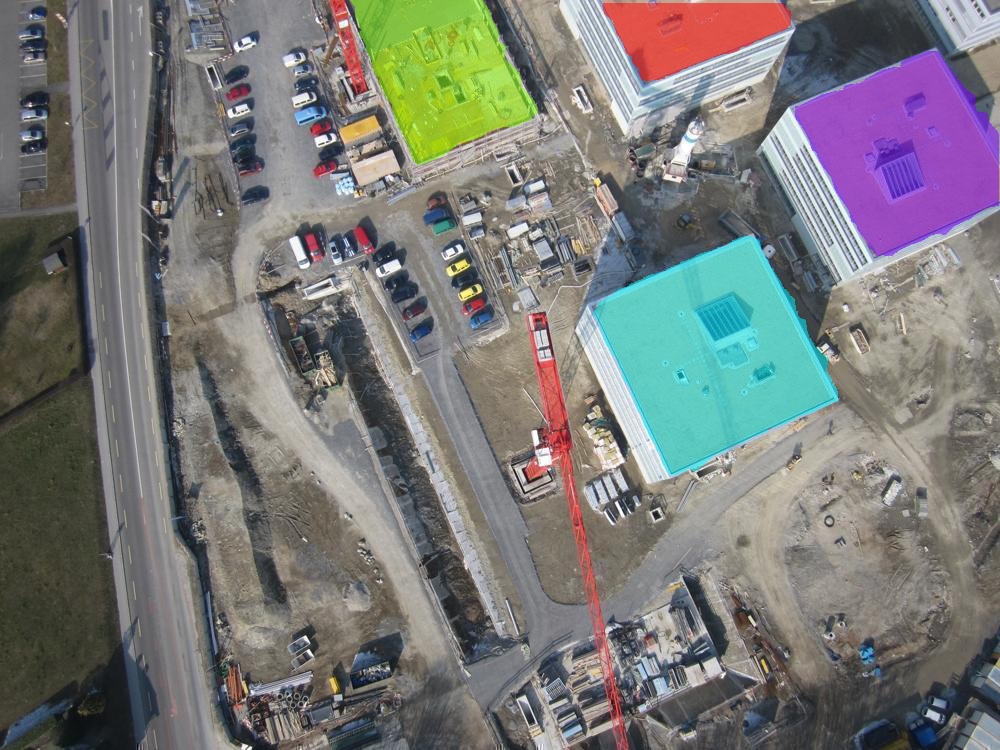} &
\includegraphics[height=1.75cm,trim=80 70 80 140,clip]{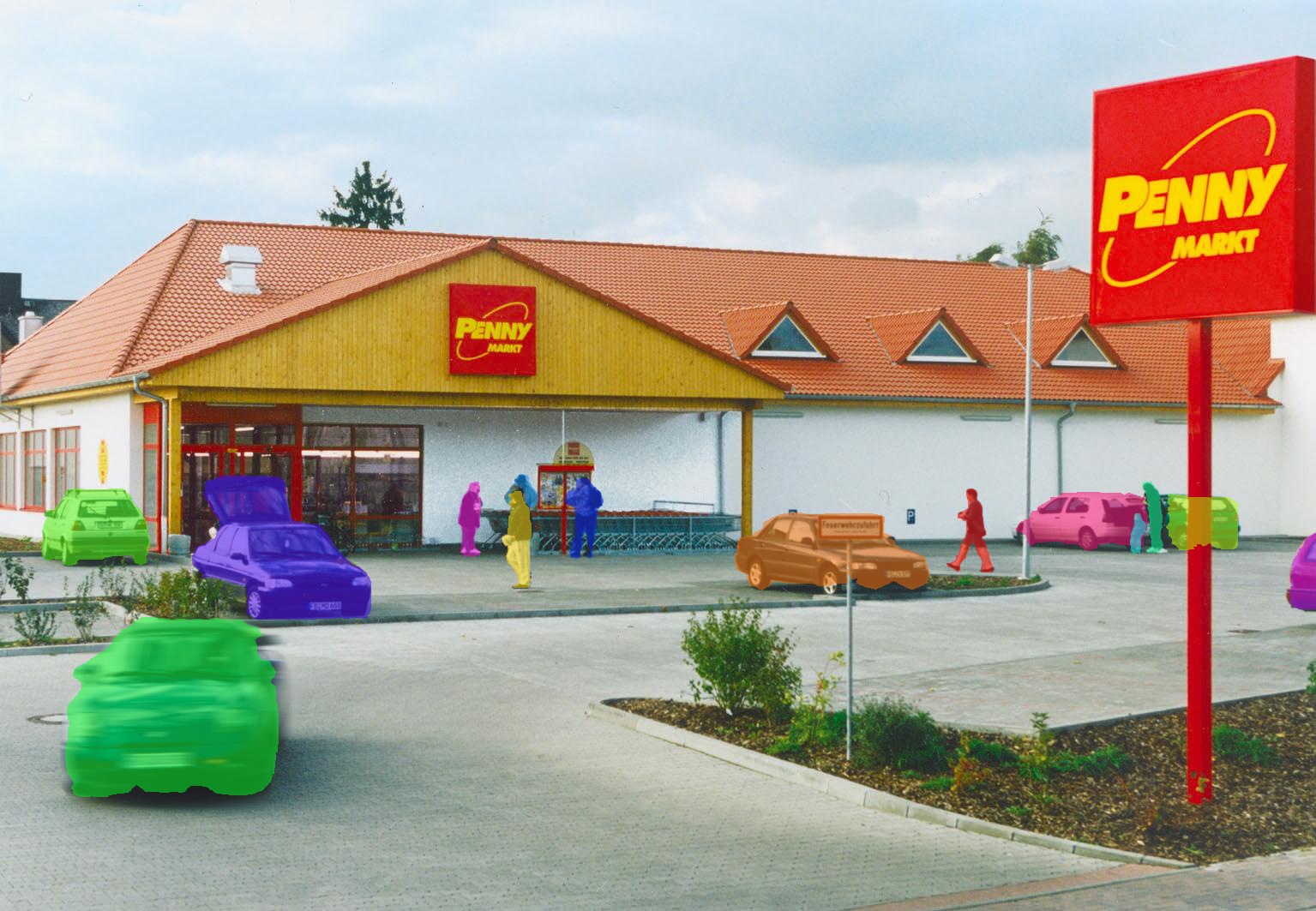}
\\[-0.5mm]
\includegraphics[height=1.75cm,trim=10 20 10 40,clip]{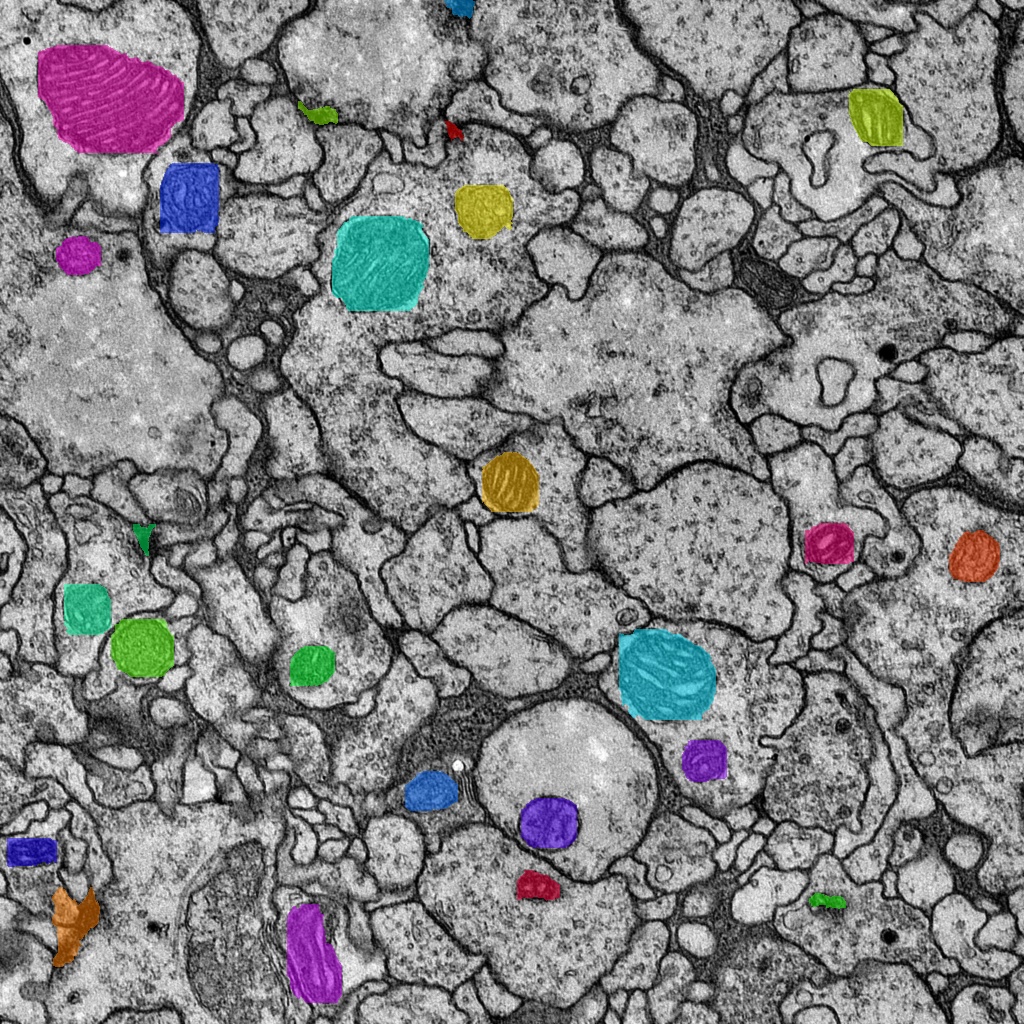} &
\includegraphics[height=1.75cm,trim=100 0 300 80,clip]{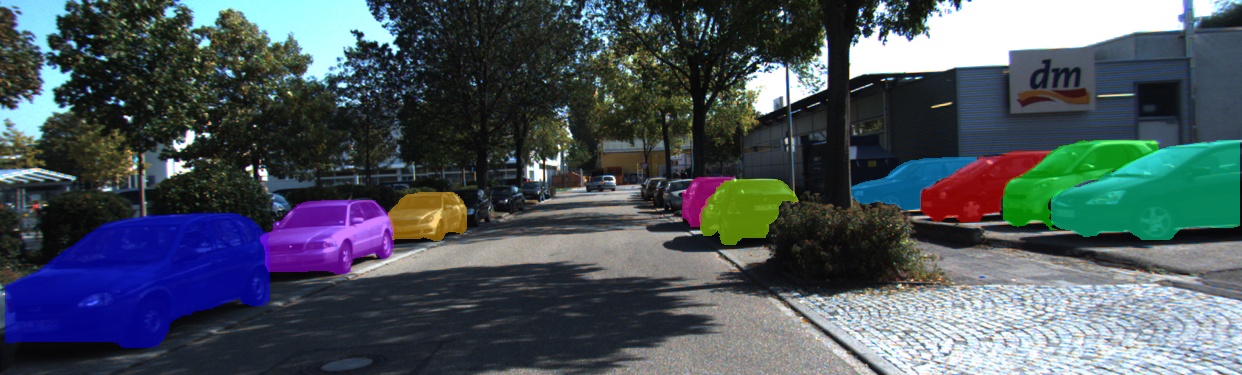} &
\includegraphics[height=1.75cm,trim=0 0 0 30,clip]{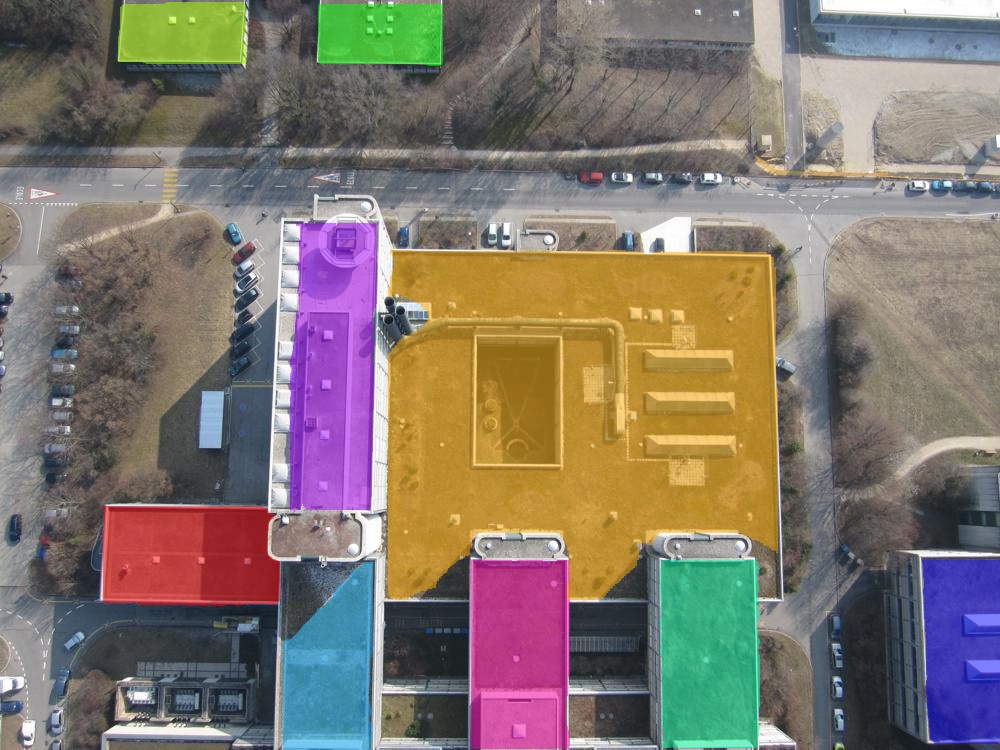} &
\includegraphics[height=1.75cm,trim=0 0 0 70,clip]{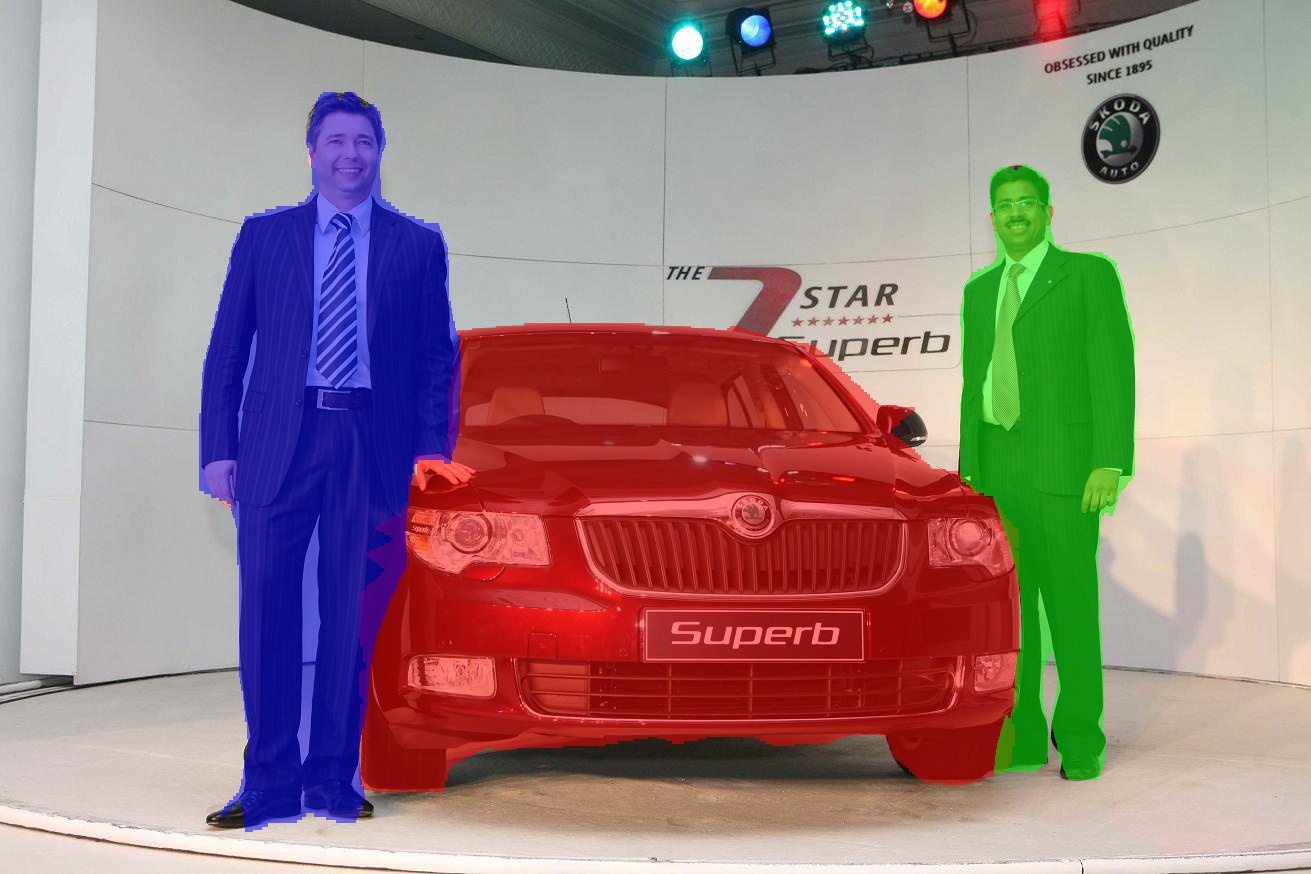} 
\end{tabular}
\caption{{\bf Qualitative results:}  Note that the method employs ground-truth boxes. The first row is from Cityscapes. In the bottom two rows, from left to right: Medical, KITTI, Rooftop, ADE.}
\label{fig:def_grid_city_ins_seg}
\end{figure*}

\subsubsection{Pixel-wise Object Instance Annotation}

\paragraph{\textbf{Network Architecture:}} To predict the class label for each deformed grid's cell, we first average the feature of all pixels inside each cell, and use a 4-layer MLP to predict the probability of foreground/background. For the hyperparameters and architecture details, we refer to the appendix. 

\paragraph{\textbf{Results:}} Table~\ref{tbl:defgrid-pixel-seg} provides quantitative results.  We show qualitative results in  the appendix. Predicting the (binary) class label over the deformed grid's cells achieves higher performance than carefully designed pixel-wise baselines, demonstrating the effectiveness of reasoning on our deformed grid. 


\subsection{Unsupervised Image Partitioning}
\label{sec:exp_superpixel}
\paragraph{\textbf{Dataset:}} Following SSN~\cite{ssn}, we train the model on 200 training images in the BSDS500~\cite{bsd} and evaluate on 200 test images. Details are provided in appendix. 

\begin{figure*}[t!]
\centering
\addtolength{\tabcolsep}{-4.2pt}
\begin{tabular}{cccc}
\includegraphics[height=2.4cm,width=0.27\linewidth,trim=0 5 0 0,clip]{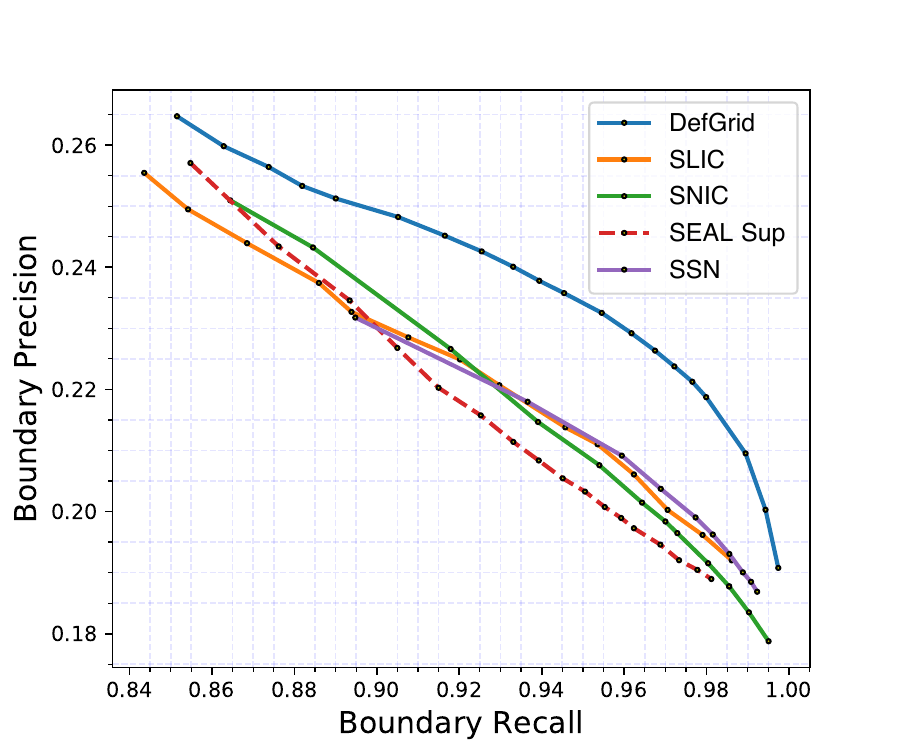} &
\includegraphics[height=2.4cm,width=0.27\linewidth,trim=0 5 0 0,clip]{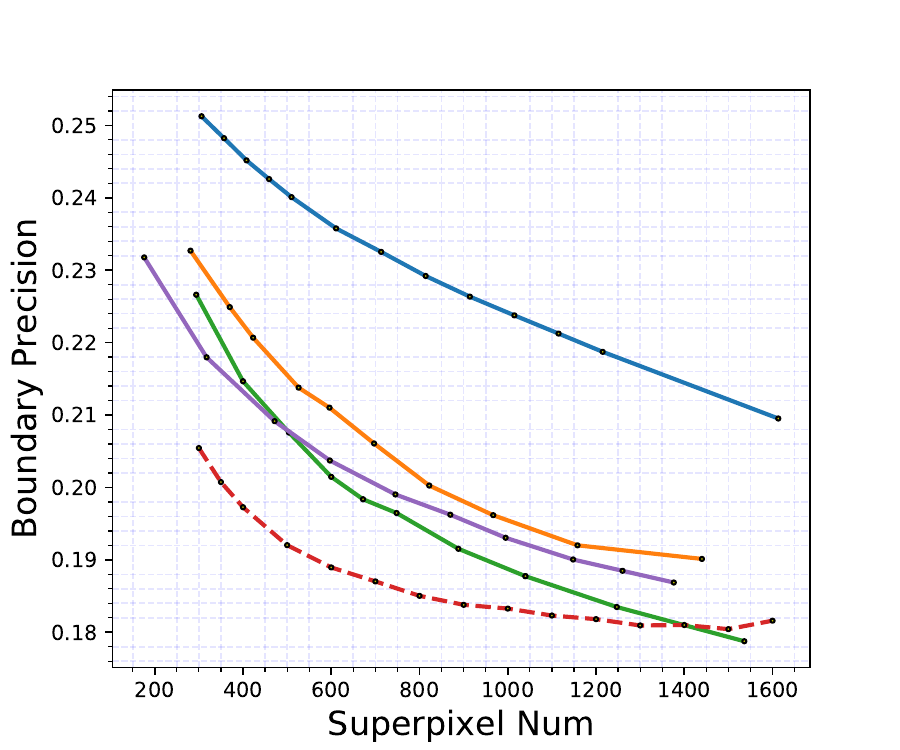} &
\includegraphics[height=2.4cm,width=0.27\linewidth,trim=0 5 0 0,clip]{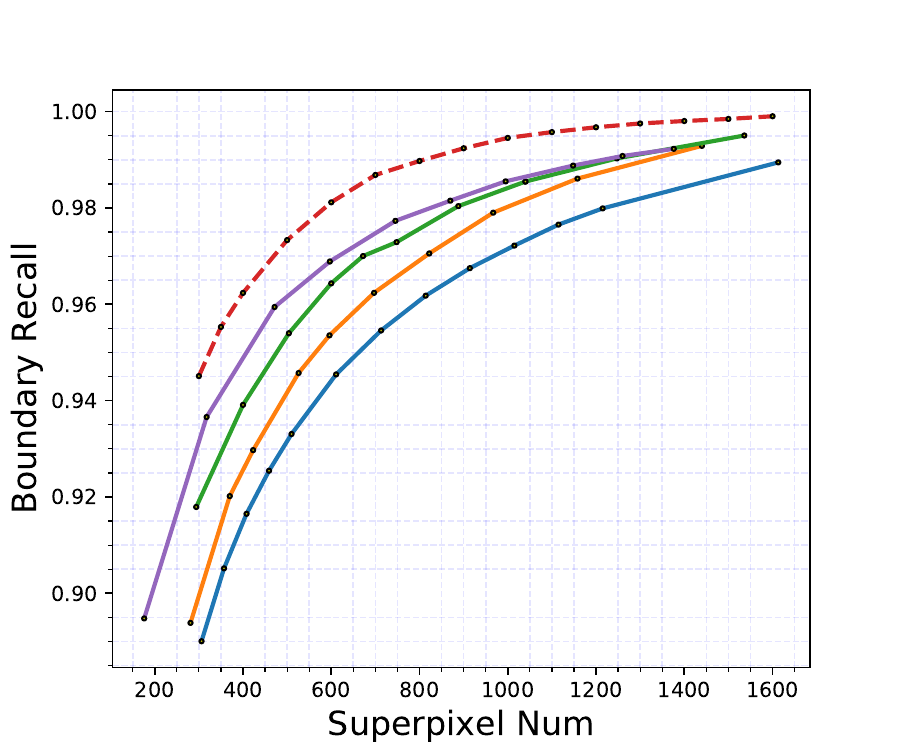} &
\includegraphics[height=2.4cm,width=0.27\linewidth,trim=0 5 0 0,clip]{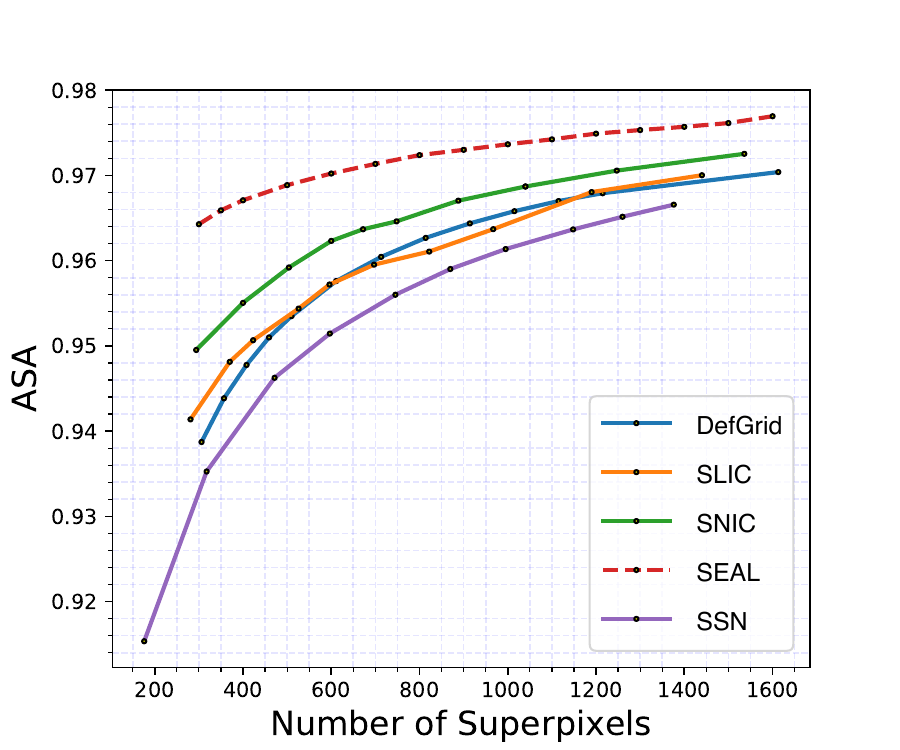}  
\end{tabular}
\caption{ {\bf Unsupervised Image Partitioning}. From left to right: BP-BR, BP, BR and ASA. We use dotted line to represent supervised method. }
\label{fig:def_grid_asa_and_br}
\end{figure*}

\begin{figure*}[t!]
\centering
\addtolength{\tabcolsep}{-0.3mm}
\begin{tabular}{ccccc}
\begin{minipage}[t] {0.197\textwidth}
 \centering
 {\scriptsize SLIC}\\[0.3mm]
\includegraphics[width=\linewidth]{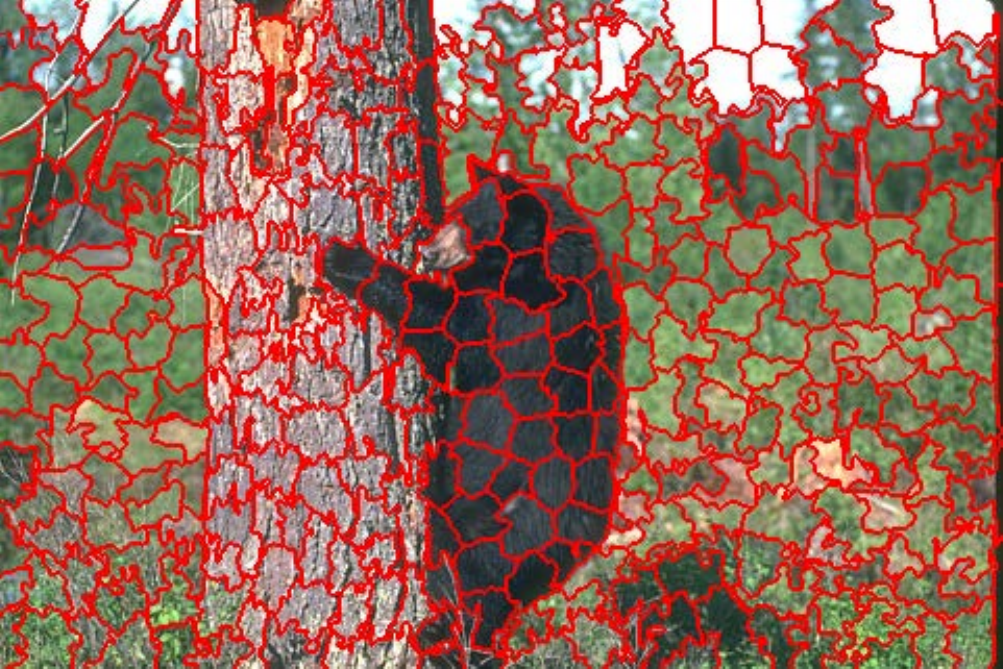} \\
\includegraphics[width=\linewidth]{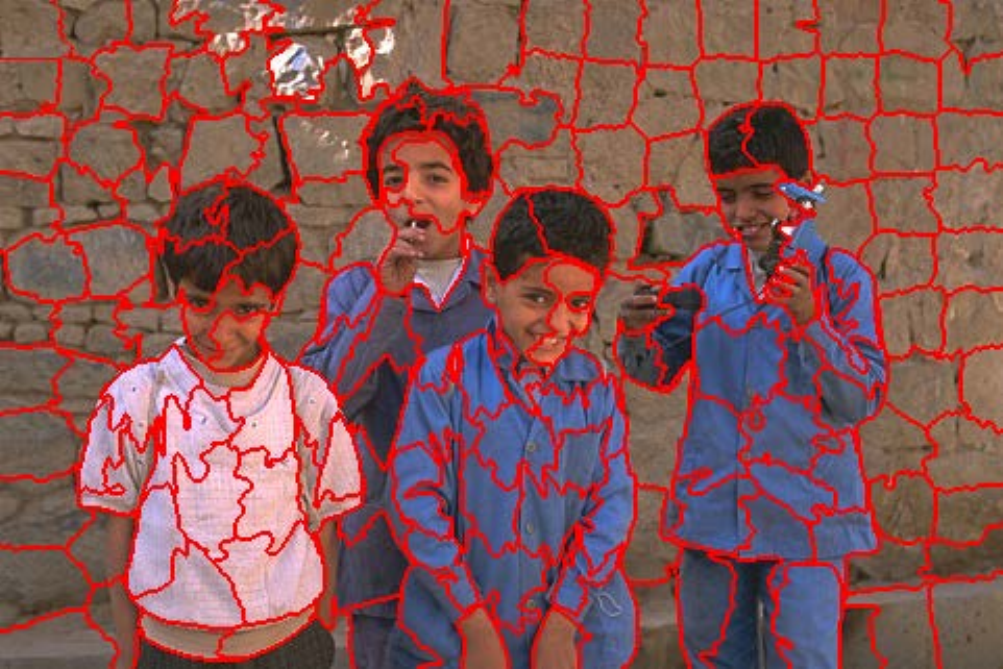} \\
\includegraphics[width=\linewidth]{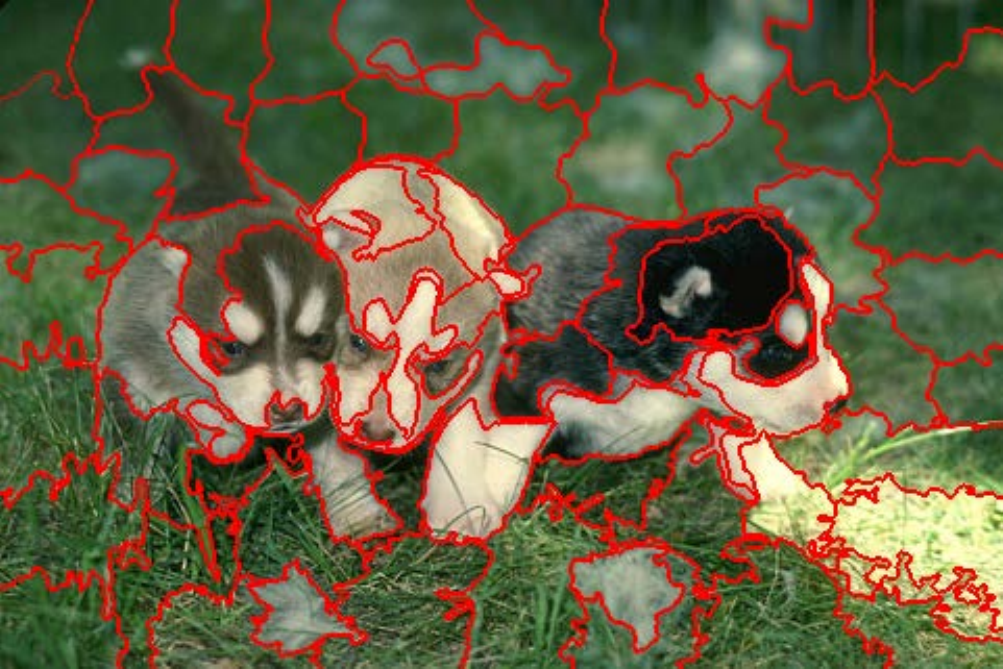} 
\end{minipage}
&
\begin{minipage}[t] {0.197\textwidth}
 \centering
 {\scriptsize SNIC}\\[0.3mm]
\includegraphics[width=\linewidth]{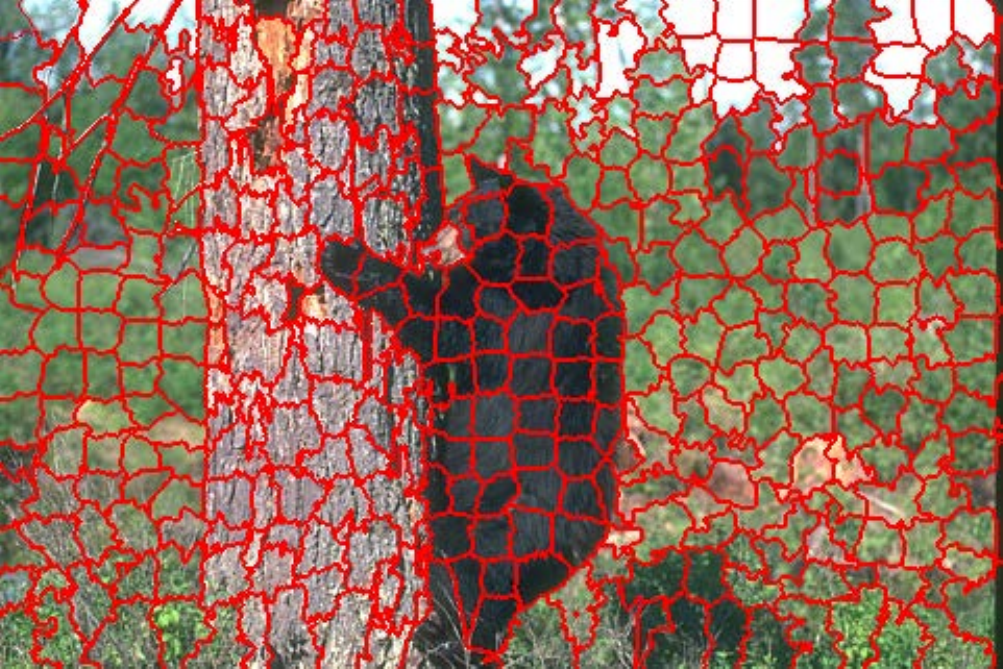} \\
\includegraphics[width=\linewidth]{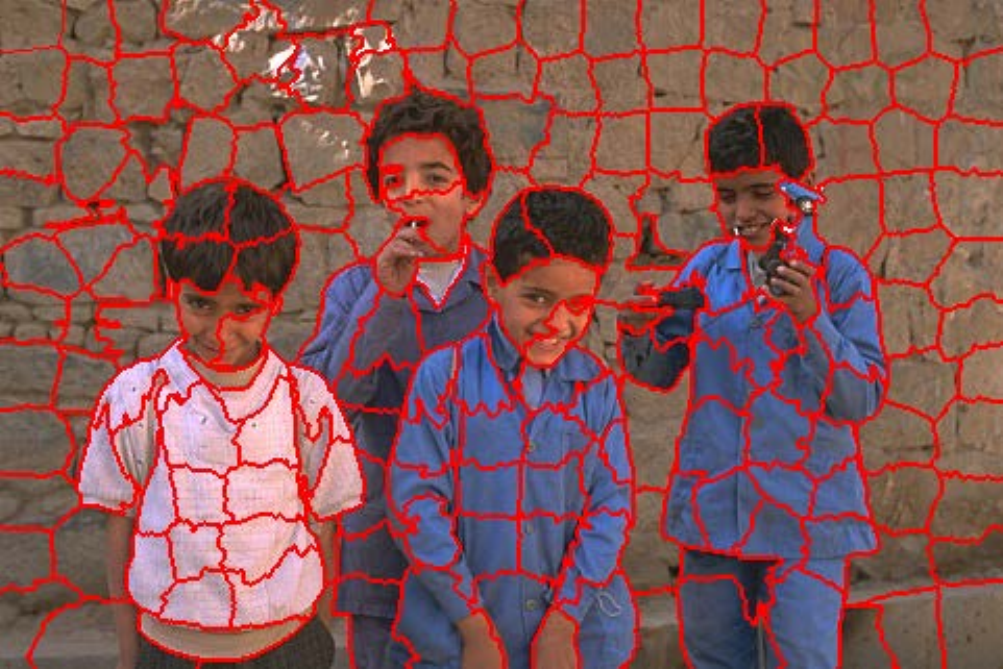} \\
\includegraphics[width=\linewidth]{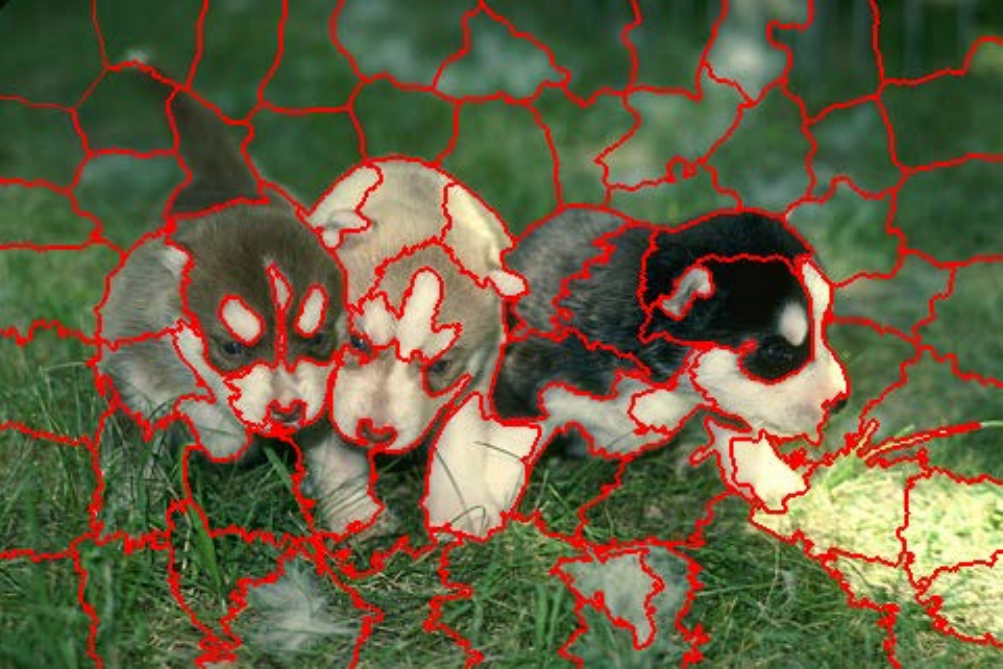} 
\end{minipage}
\begin{minipage}[t] {0.197\textwidth}
 \centering
 {\scriptsize SEAL}\\[0.3mm]
\includegraphics[width=\linewidth]{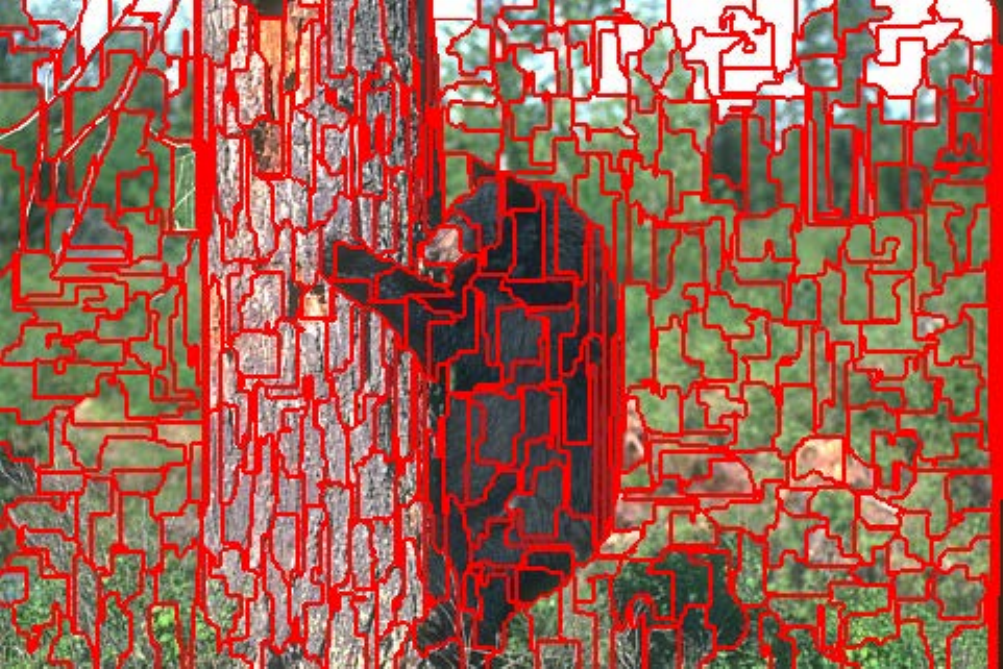} \\
\includegraphics[width=\linewidth]{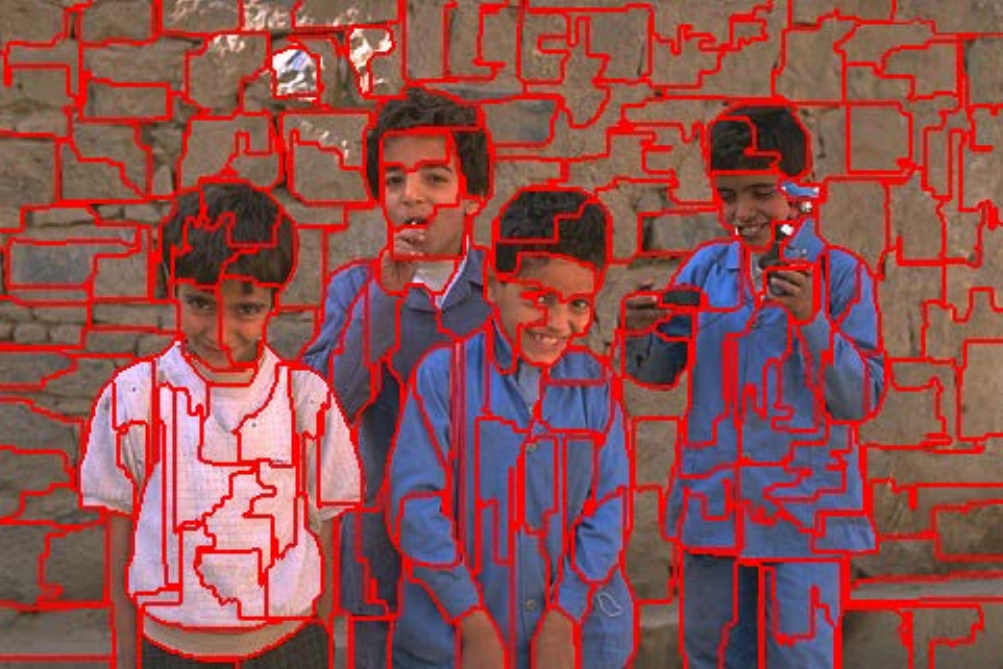} \\
\includegraphics[width=\linewidth]{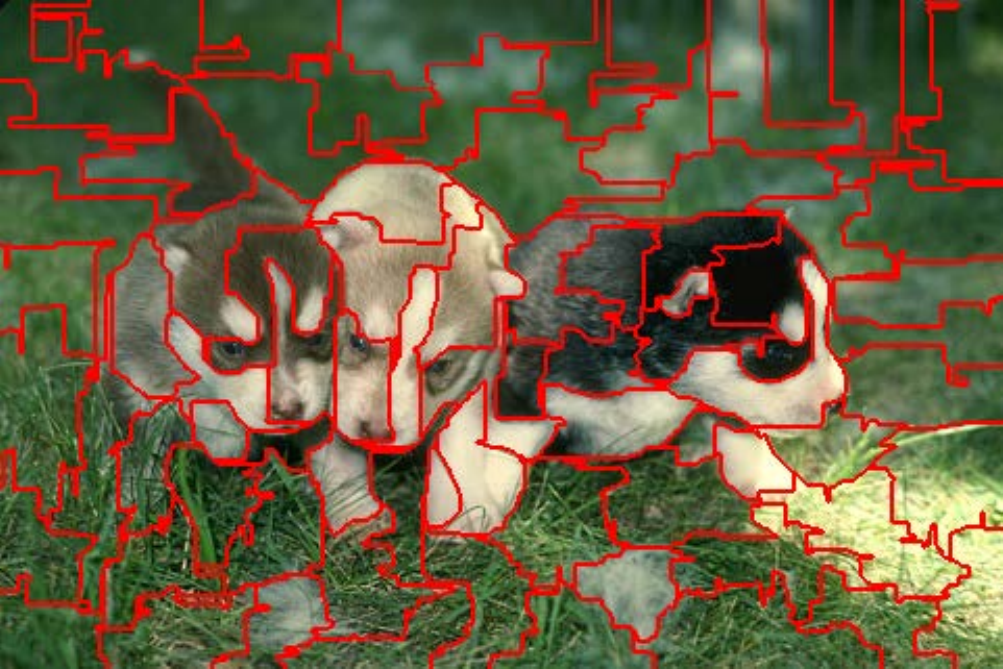} 
\end{minipage} 
&
\begin{minipage}[t] {0.197\textwidth}
 \centering
 {\scriptsize {\ourmodel} }\\[0.3mm]
\includegraphics[width=\linewidth, trim=70 62 60 63,clip]{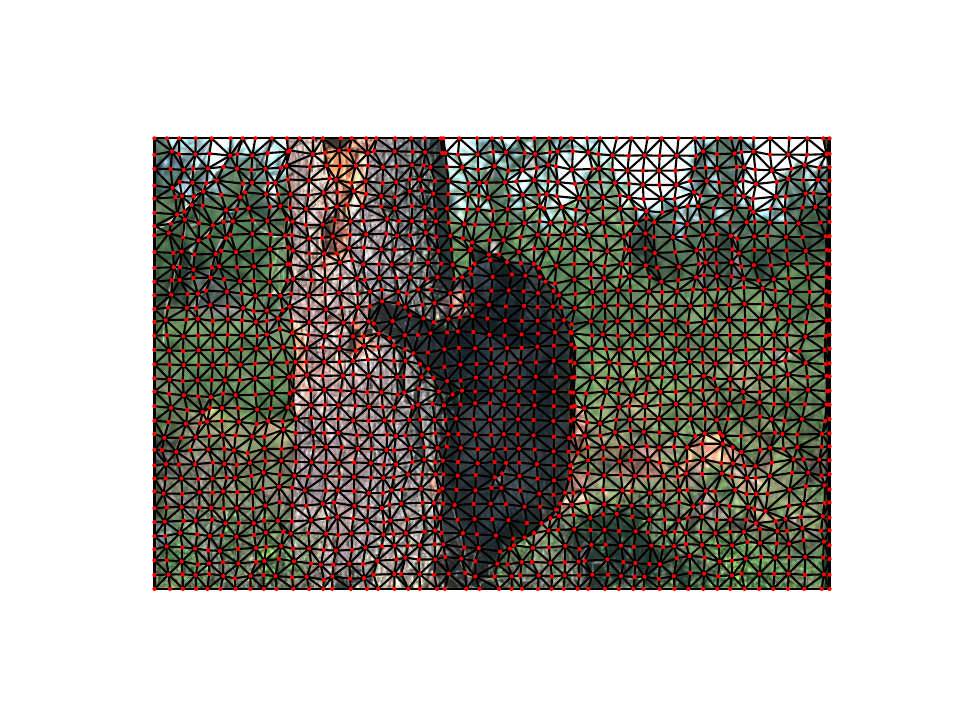} \\
\includegraphics[width=\linewidth, trim=70 62 60 63,clip]{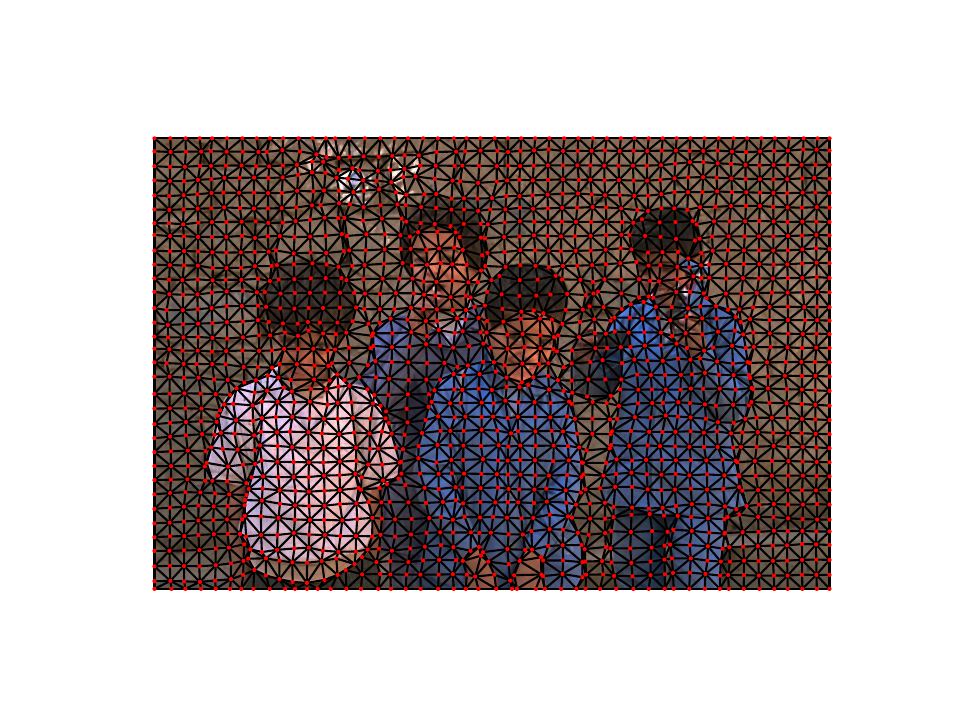} \\
\includegraphics[width=\linewidth,trim=70 62 60 63,clip]{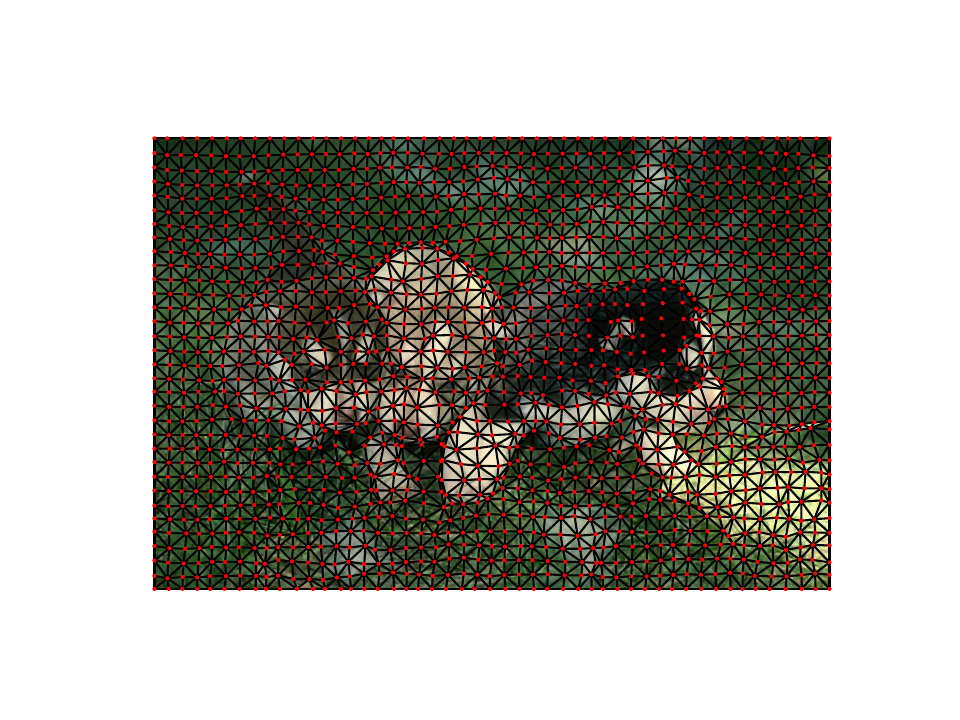} 
\end{minipage}
&
\begin{minipage}[t] {0.197\textwidth}
 \centering
 {\scriptsize {\ourmodel} -- Merging}\\[0.3mm]
\includegraphics[width=\linewidth]{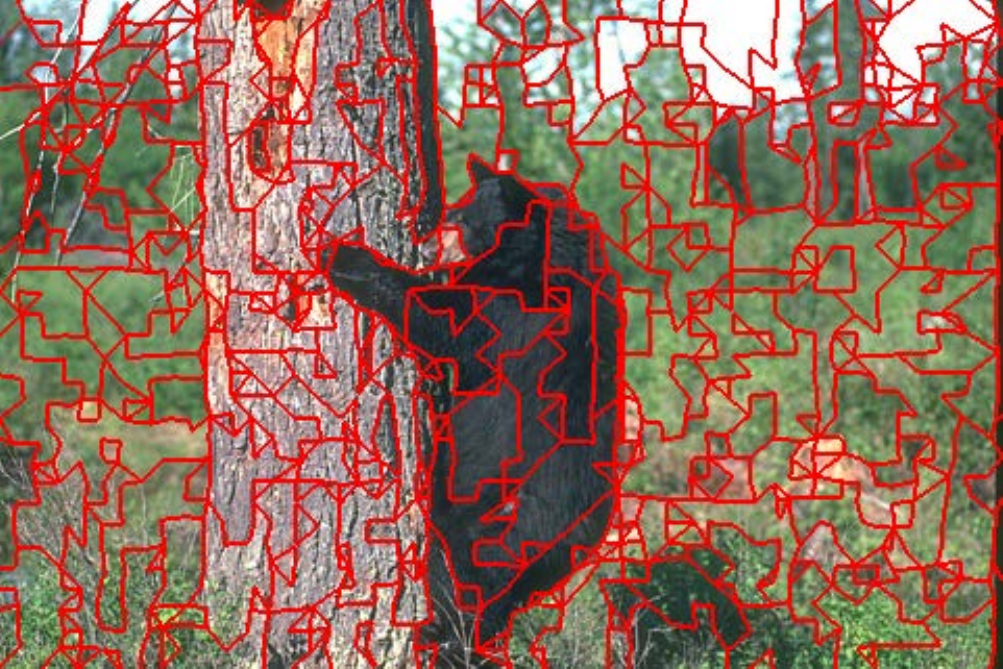} \\
\includegraphics[width=\linewidth]{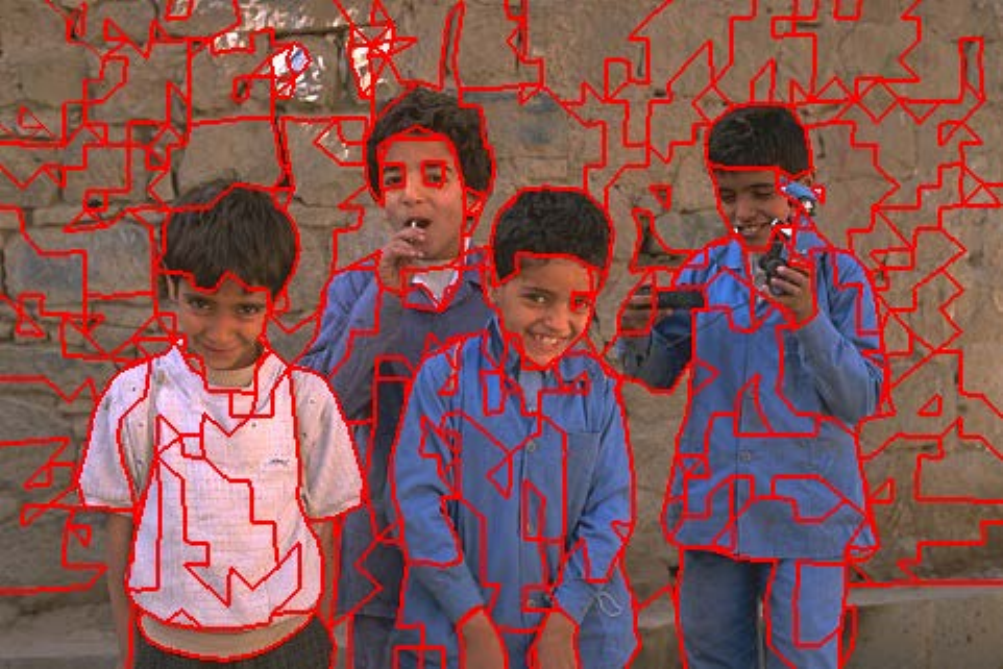} \\
\includegraphics[width=\linewidth]{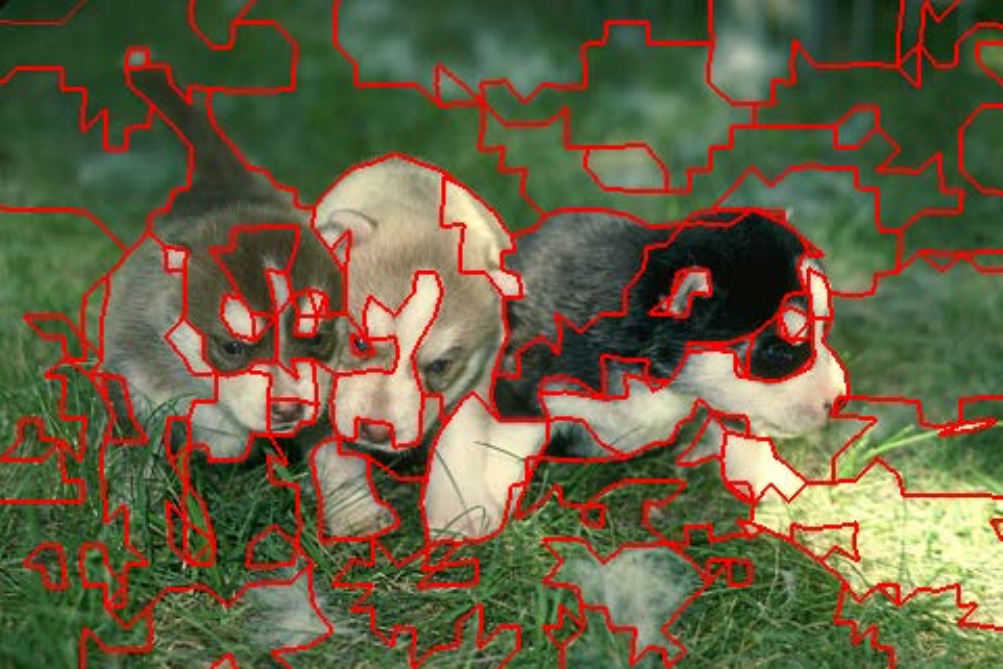} 
\end{minipage}

\end{tabular}
\caption{{\bf Unsupervised Image Partitioning.} We compare the {\ourmodel} and results after clustering with existing superpixel baselines. [Please zoom in]}
\label{fig:def_grid_superpixel_examples}
\end{figure*}


\paragraph{\textbf{Evaluation Metric:}} Following the SSN and SEAL~\cite{ssn,seal}, we use Achievable Segmentation Accuracy (ASA), Boundary Precision (BP) and Boundary Recall (BR) to evaluate the performance of superpixels. For BP and BR, we set the tolerance to be 3 pixels. The evaluation scripts are from SEAL\footnote{https://github.com/wctu/SEAL}.

\paragraph{\textbf{Baselines:}} We compare our method with both traditional superpixel methods, SLIC~\cite{slic}, SNIC~\cite{snic}, and deep learning based method SSN~\cite{ssn}, SEAL~\cite{seal}. Note that SEAL not only utilizes ground-truth annotation for training, but also is trained on validation set. We use the official codebase and trained model provided by the authors. We perform all comparisons at different numbers of superpixels. 

\paragraph{\textbf{Results:}} Quantitative and qualitative results are presented in  Fig~\ref{fig:def_grid_asa_and_br}, Fig~\ref{fig:def_grid_superpixel_examples}, respectively. 
The deformed grid aligns well with object boundary and outperforms all unsupervised baselines in terms of BP and BP-BR curve, with comparable performance with counterparts in terms of ASA and BR. Our intuition is that, since the edge between two vertices in our grid is constrained to be a straight line, while the ground truth annotation is labelled pixel-by-pixel, our grid sacrifices a little boundary recall while achieving higher boundary precision.
It is worth to note that our method outperforms the supervised method SEAL~\cite{seal}. This reflects the fact that an appearance feature provides a useful signal for training our {\ourmodel} , and our method effectively utilizes this signal.
\vspace{-4pt}
\section{Conclusion}
In this paper, we proposed  to deform a regular grid to better align with image boundaries as a more efficient way to process images. Our {\ourmodel} is a neural network that predicts offsets for vertices in a grid to perform the alignment, and can be trained entirely with unsupervised losses. We showcase our approach in several downstream tasks  with significant performance gains. Our method produces accurate superpixel segmentations, is significantly more precise in outlining object boundaries particularly on out of domain datasets, and leads to a large improvements for semantic segmentation. We hope that our {\ourmodel} benefits other computer vision tasks  when combined with deep networks.

\textbf{Acknowledgments:} This work was supported by NSERC. SF acknowledges the Canada CIFAR AI Chair award at the Vector Institute. 
We thank Frank Shen, Wenzheng Chen and Huan Ling for helpful discussions, and also thank the anonymous reviewers for valuable comments. 


\clearpage

\bibliographystyle{splncs04}

\end{document}